\PassOptionsToPackage{backref,pagebackref}{hyperref}
\documentclass[11pt]{article}

\usepackage[top=30truemm,bottom=30truemm,left=25truemm,right=25truemm]{geometry}

\usepackage[utf8]{inputenc} 
\usepackage[T1]{fontenc}    
\usepackage{textcomp}
\usepackage{bm}
\usepackage{bbm}  

\usepackage{amsmath}
\usepackage{amssymb}
\usepackage{amsthm}
\usepackage{amsfonts}
\usepackage{mathrsfs}
\usepackage{nicefrac}
\usepackage{mathtools}

\usepackage{booktabs}
\usepackage{multirow}

\usepackage[authoryear,round]{natbib}

\usepackage[svgnames]{xcolor}
\usepackage{colortbl}  
\usepackage[hypertexnames=false]{hyperref}       
\hypersetup{
  colorlinks=true,
  linkcolor=blue,
  citecolor=blue,
}

\usepackage{stmaryrd}
\usepackage{tgtermes}
\usepackage{comment}

\usepackage[ruled,vlined,linesnumbered]{algorithm2e}
\SetArgSty{textnormal}

\usepackage{microtype}
\usepackage[labelfont=bf]{caption}
\usepackage{wrapfig}
\usepackage{tikz}

\usepackage{url}

\usepackage{enumitem}

\usepackage{thm-restate}

\usepackage[normalem]{ulem}
\usepackage{tcolorbox}
\usepackage{mdframed} 

\usepackage{threeparttable}

\usepackage[nameinlink]{cleveref}
\crefname{equation}{Eq.}{Eqs.}
\Crefname{algocf}{Algorithm}{Algorithms}  

\theoremstyle{plain}
\newtheorem{theorem}{Theorem}

\newtheorem{lemma}[theorem]{Lemma}

\theoremstyle{definition}

\renewcommand{\hat}{\widehat}
\renewcommand{\tilde}{\widetilde}
\renewcommand{\epsilon}{\varepsilon}

\def\e{\mathrm{e}}

\def\R{\mathbb{R}}
\def\Rp{\mathbb{R}_{> 0}} 
 
\def\N{\mathbb{N}}

\def\calK{\mathcal{K}}
\def\calL{\mathcal{L}}

\DeclareMathOperator*{\argmin}{arg\,min}

\DeclareMathOperator{\dom}{dom}

\DeclarePairedDelimiter{\brk}{[}{]}

\DeclarePairedDelimiter{\set}{\{}{\}}
\DeclarePairedDelimiter{\prn}{(}{)}
\DeclarePairedDelimiter{\nrm}{\|}{\|}

\DeclarePairedDelimiter{\inpr}{\langle}{\rangle}  

\renewcommand{\d}{\mathrm{d}}

\newcommand{\zeros}{\mathbf{0}}
\newcommand{\ones}{\mathbf{1}}

\newcommand{\ie}{\textit{i.e.,}}
\newcommand{\eg}{\textit{e.g.,}}

\newcommand{\Reg}{\mathsf{Reg}}
\newcommand{\SReg}{\mathsf{SocialReg}}
\newcommand{\DReg}{\mathsf{DReg}}

\newcommand{\xhat}{\widehat{x}}
\newcommand{\yhat}{\widehat{y}}
\newcommand{\lhat}{\widehat{\ell}}
\newcommand{\ghat}{\widehat{g}}

\newcommand{\nn}{\nonumber\\}
\newcommand{\n}{\nonumber}
\newcommand{\per}{\,.}
\newcommand{\com}{\,,}

\newcommand{\sumT}{\sum_{t=1}^T}

\title{Tight Regret Upper and Lower Bounds for Optimistic Hedge \\ in Two-Player Zero-Sum Games}

\author{
  Taira Tsuchiya\footnote{
    The University of Tokyo and RIKEN; 
    \texttt{tsuchiya@mist.i.u-tokyo.ac.jp}.
  }
}

\begin{document}
\maketitle

\begin{abstract}
In two-player zero-sum games, the learning dynamic based on optimistic Hedge achieves one of the best-known regret upper bounds among strongly-uncoupled learning dynamics. With an appropriately chosen learning rate, the social and individual regrets can be bounded by $O(\log(mn))$ in terms of the numbers of actions $m$ and $n$ of the two players. This study investigates the optimality of the dependence on $m$ and $n$ in the regret of optimistic Hedge. To this end, we begin by refining existing regret analysis and show that, in the strongly-uncoupled setting where the opponent's number of actions is known, both the social and individual regret bounds can be improved to $O(\sqrt{\log m \log n})$. In this analysis, we express the regret upper bound as an optimization problem with respect to the learning rates and the coefficients of certain negative terms, enabling refined analysis of the leading constants. We then show that the existing social regret bound as well as these new social and individual regret upper bounds cannot be further improved for optimistic Hedge by providing algorithm-dependent individual regret lower bounds. Importantly, these social regret upper and lower bounds match exactly including the constant factor in the leading term. Finally, building on these results, we improve the last-iterate convergence rate and the dynamic regret of a learning dynamic based on optimistic Hedge, and complement these bounds with algorithm-dependent dynamic regret lower bounds that match the improved bounds.
\end{abstract}
\section{Introduction}\label{sec:introduction}

\begin{table*}[t]
      \caption{Regret upper and lower bounds of learning dynamics based on optimistic Hedge for the $x$-player in two-player zero-sum games.
        The lower bounds are algorithm-dependent and correspond to the learning rates used in the upper bounds.
        We write $M = \log m$ and $N = \log n$.
        The individual regret upper bounds are for the case where the focus is solely on minimizing the regret of the $x$-player.
        The upper bounds when minimizing the maximum of the individual regrets are provided in \Cref{thm:optimal_max_indivreg_su,thm:optimal_max_indivreg_casu}.
        The learning rates corresponding to each regret bound are summarized in \Cref{table:lr} in \Cref{app:experiments}.
    }
    \label{table:regret}
    \centering
    \resizebox{\textwidth}{!}{
    \begin{tabular}{lll}
      \toprule
      & Upper bound & Lower bound
      \\
      \midrule
      \multicolumn{2}{l}{{\normalsize $\triangleright$ \uline{Strongly-uncoupled learning dynamics} }} & 
      \vspace{2pt}
      \\
      Social regret & $2 (M + N) + 1$ \ \citep{rakhlin13optimization} & $2 (M + N) - o(1)$ \ (\textbf{This work, \Cref{thm:lower_bounds}})
      \\
      Individual regret & $M + N + 1/2$ \ \citep{rakhlin13optimization}, \Cref{eq:x_best_indiv_su} & $M - o(1)$ \ (\textbf{This work, \Cref{thm:lower_bounds}})
      \\
      Dynamic regret & $(2 (M + N) + 1) \log T$ \ \citep{cai25from} & $M \log (T + 1) - o(1)$ \ (\textbf{This work, \Cref{thm:lower_bounds_dynamic}})
      \\  
      \midrule
      \multicolumn{2}{l}{{\normalsize $\triangleright$ \uline{Cardinality-aware strongly-uncoupled learning dynamics}}} & 
      \vspace{2pt}
      \\
      Social regret & $2 \sqrt{M (N + 1/2)} + 2 \sqrt{N (M + 1/2)}$ \ (\textbf{This work, \Cref{thm:optimal_sreg_casu}}) & $\simeq 2 \sqrt{M N}  - o(1)$ \ (\textbf{This work, \Cref{thm:lower_bounds}})
      \\
      Individual regret & $2 \sqrt{M (N + 1/2)}$ \ (\textbf{This work, \Cref{eq:x_best_indiv_casu}}) & $\simeq \sqrt{M N} - o(1)$ \ (\textbf{This work, \Cref{thm:lower_bounds}})
      \\
      Dynamic regret & $2 \prn{\sqrt{M (N + 1/2)} + \sqrt{N (M + 1/2)}} \log T$ \ (\textbf{This work, \Cref{thm:dynamic_reg_casu}}) & $\simeq \sqrt{M N} \log T - o(1)$ \ (\textbf{This work, \Cref{thm:lower_bounds_dynamic}})
      \\  
      \bottomrule
    \end{tabular}
    }
  \end{table*}

Learning in games is a central problem in both game theory and machine learning, and it is well known that players can learn an equilibrium by employing online learning algorithms~\citep{freund99adaptive,hart00simple,cesabianchi06prediction}.
Such equilibrium learning has led to the development of AI systems that surpass human performance~\citep{bowling15heads,moravvcik17deepstack,perolat22mastering,meta22human}.
Furthermore, its effectiveness has also been demonstrated recently for the alignment of large language models (LLMs)~\citep{munos24nash,swamy24minimaxlist}.

The advantage of equilibrium learning based on online learning is that it can be realized through \emph{uncoupled} learning dynamics~\citep{hart03uncoupled}.
In particular, in two-player zero-sum games, it can be achieved by \emph{strongly-uncoupled} learning dynamics~\citep{daskalakis11near}, namely dynamics in which each player learns solely from the rewards they have observed in the past, without knowing the opponent's strategies, observations, or even the number of actions.
Such dynamics, which attain an approximate equilibrium as a consequence of maximizing cumulative reward based only on their own observations, are consistent with realistic behavioral models~\citep{hart03uncoupled},
and many recent learning dynamics based on online learning exhibit this property (\eg~\citealt{syrgkanis15fast,anagnostides22uncoupled}).

In learning in games, the most representative online learning algorithm adopted by each player is the Hedge algorithm~\citep{littlestone94weighted,freund97decision}.
The Hedge algorithm selects actions using weights exponentially scaled by past cumulative rewards, and guarantees a worst-case (external) regret of $O(\sqrt{T \log m})$, where $T$ is the number of rounds and $m$ is the number of actions.

In learning in games, this worst-case regret upper bound can be significantly improved if each player employs specific online learning algorithms.
In particular, in two-player zero-sum games, one of the most powerful algorithms is \emph{optimistic Hedge}~\citep{rakhlin13optimization,syrgkanis15fast}, which selects actions according to weights that are exponentially scaled not only by the cumulative rewards but also by the most recently observed reward.
When the learning rates of optimistic Hedge are set to an absolute constant, the social regret, \ie~the sum of the regrets of all players, can be bounded by~$O(\log (m n))$, where $m$ and $n$ denote the numbers of actions of the $x$- and $y$-players, respectively.
This implies convergence to a Nash equilibrium at the rate of $O(\log (m n) / T)$, which is optimal up to the $\log (m n)$ factor~\citep{daskalakis11near}.

While using standard no-regret online learning algorithms such as optimistic Hedge, one typically guarantees only average-iterate (time-averaged) convergence rather than last-iterate convergence,
very recent work shows that, by employing a learning dynamic that outputs the time-averaged strategy of optimistic Hedge, one can also guarantee (anytime) last-iterate convergence~\citep{cai25from}.
Specifically, this approach achieves $O(\log (m n) / t)$ last-iterate convergence at every round $t$, which implies that each player's dynamic regret is bounded by $O(\log (m n) \log T)$.

As we have seen, optimistic Hedge is one of the best algorithms for learning dynamics in two-player zero-sum games.
However, even with this learning dynamic, there remains a gap of an $O(\log (m n))$ factor compared with the existing lower bound on the convergence rate.
This raises the fundamental question: what are the optimal upper bounds for the social and individual regrets when using uncoupled learning dynamics?
Despite its fundamental nature, this question has not yet been investigated.
Additional related work that could not be included in the main text is provided in \Cref{sec:related_work}.

\paragraph{Contributions of This Paper}
As a first step toward addressing this open question, this study investigates the following question:
in learning two-player zero-sum games, how optimal are optimistic-Hedge-based learning dynamics and their analysis in terms of dependence on the numbers of actions $m$ and $n$ and on the leading constants?
To answer this question, we make the following contributions.

As a first step, in \Cref{sec:regret_upper_bounds}, we begin by refining the existing regret analysis of optimistic Hedge so that we can compare it precisely with the lower bounds we derive.
The existing upper bounds are somewhat ad hoc: the analysis pays little attention to the magnitude of the leading constants, and although exploiting a certain negative term that appear in the upper bound of optimistic Hedge is crucial, it has not been treated with sufficient care.
We therefore conduct a careful analysis to investigate how much we can improve the leading constants of the regret bounds and their dependence on $m$ and $n$.

In analyzing optimistic Hedge, in addition to tuning the learning rate, it is important to exploit the negative term.
In our analysis, we observe that this negative term plays two distinct roles and introduce a new parameter to capture the tradeoff between them.
We then express the regret upper bound as an optimization problem.
This formulation elucidates the tradeoff between the $x$- and $y$-players' individual regrets, which allows us to make a precise comparison with the individual regret lower bound derived next.
Using this optimization perspective, we show that, in strongly-uncoupled learning dynamics where each player is additionally allowed to know the opponent's number of actions, one can achieve social and individual regret bounds of $O(\sqrt{\log m \log n})$.
This improvement is particularly effective in games where $\log m$ and $\log n$ are highly imbalanced.
In particular, this occurs when the number of actions of a player is exponentially large: for example, network interdiction, where the set of source--sink paths is exponential~\citep{washburn95two}; extensive-form games whose normal-form strategy space is exponential~\citep{koller96efficient}; zero-sum games with submodular structure~\citep{wilder18equilibrium}; and asymmetric combinatorial--continuous zero-sum games~\citep{li25can}.
Through numerical experiments, we confirm that being cardinality-aware indeed leads to empirical improvements in both the social regret and the maximum of the individual regrets.
The detailed experimental setup and results are provided in \Cref{app:experiments}.

Next, in \Cref{sec:lower_bound}, to investigate the optimality of the refined regret upper bounds, we derive regret lower bounds for the learning dynamic based on optimistic Hedge.
We show that when each player uses optimistic Hedge with learning rates $\eta, \eta' > 0$, their individual regrets are lower bounded by $\log (m) / \eta - o(1)$ and $\log (n) / \eta' - o(1)$, respectively, where $o(1)$ denotes a term that vanishes as $T \to \infty$.
These regret lower bounds imply that the social regret of optimistic Hedge matches the existing best social regret bounds \emph{including leading constants}.
For the individual regret, the upper and lower bounds match in many cases up to constant factors.
To our knowledge, our work is the first to derive regret lower bounds for optimistic Hedge and to investigate their dependence on the numbers of actions $m$ and $n$.

As the third contribution, in \Cref{sec:dynamic}, we extend the above refinement and analysis of the (external) regret upper and lower bounds to dynamic regret.
First, we present an improved result on the convergence rate of the last iterate (and the corresponding dynamic regret bound) for learning dynamics based on optimistic Hedge that enjoy the last-iterate convergence property.
We then provide an improvement of the dynamic regret itself, together with an algorithm-dependent dynamic regret lower bound that matches these results.
From these derived lower bounds, we can see that the approach of \citet{cai25from} cannot be further improved with respect to $m$, $n$, and $T$.
These contributions are summarized in \Cref{table:regret}.

\section{Preliminaries}\label{sec:preliminaries}
This section provides some preliminaries.
For $n \in \N$, we denote $[n] = \set{1, \ldots, n}$.
We use $\zeros$ and $\ones$ to denote the all-zero and all-one vectors, respectively.
For a vector $x$, we write $x(i)$ for its $i$-th coordinate, and $\nrm{x}_p$ for its $\ell_p$-norm, where $p \in [1,\infty]$.

\subsection{Learning in Two-Player Zero-Sum Games}

\paragraph{Setup}
Learning in a two-player zero-sum game is characterized by a payoff matrix $A \in [-1, 1]^{m \times n}$,
where $m$ and $n$ denote the number of actions of the $x$- and $y$-players, respectively.
The procedure of this game is as follows:
at each round $t = 1, \dots, T$, 
the $x$-player selects a strategy $x_t \in \Delta_{m}$ and the $y$-player selects $y_t \in \Delta_n$.
Then, the $x$-player observes an expected gain vector $g_t = A y_t$ and the $y$-player observes an expected loss vector $\ell_t = A^\top x_t$.
Finally, the $x$-player gains a payoff of $\inpr{x_t, g_t}$ and the $y$-player incurs a loss of $\inpr{y_t, \ell_t}$.

The goal of each player is to minimize the (external) regret given by
$
\Reg_T^x
=
\max_{x^* \in \Delta_{m}} {\Reg_T^x(x^*)}
$
and 
$
  \Reg_T^y
  = 
  \max_{y^* \in \Delta_{n}} 
  {\Reg_T^y(y^*)}
$
for 
$
  \Reg_T^x(x^*)
  = 
  \sumT \inpr{x^* - x_t, A y_t}
  =
  \sumT \inpr{x^* - x_t, g_t}
$
and
$
  \Reg_T^y(y^*)
  = 
  \sumT \inpr{y_t - y^*, A^\top x_t}
  =
  \sumT \inpr{y_t - y^*, \ell_t}
$.
Note that, in the external regret, one compares against the strategy that maximizes the cumulative gain or the strategy that minimizes the cumulative loss.
The social regret is defined as the sum of the regrets of both players, that is,
$\SReg_T = \Reg_T^x + \Reg_T^y$.
In addition, in this paper, we also consider the following notion of dynamic regret, which compares against the best strategy at each round:
$
\DReg_T^x
=
\sumT \max_{x_t^* \in \Delta_m} \inpr{x_t^* - x_t, g_t}
$
and
$
\DReg_T^y
=
\sumT \max_{y_t^* \in \Delta_n} \inpr{y_t - y_t^*, \ell_t}
\per
$

\paragraph{No-regret learning and Nash equilibrium}
We say that a pair of probability distributions $(x^*, y^*)$ over action sets $[m]$ and $[n]$ is an \emph{$\epsilon$-approximate Nash equilibrium} for $\epsilon \geq 0$ if for any distributions $x \in \Delta_m$ and $y \in \Delta_n$,
  $
    \inpr{x, A y^*} - \epsilon
    \leq
    \inpr{x^*, A y^*}
    \leq 
    \inpr{x^*, A y} + \epsilon
    .
  $
  The pair $(x^*, y^*)$ is a \emph{Nash equilibrium} if it is a $0$-approximate Nash equilibrium.

It is well known that an approximate Nash equilibrium is obtained as a consequence of no-regret learning dynamics:
\begin{theorem}[{\citealt{freund99adaptive}}]\label{thm:sreg_to_nash}
Let the sequences of strategies $(x_t)_{t=1}^T$ and $(y_t)_{t=1}^T$ be generated by online learning algorithms with regrets $\Reg_T^x$ and $\Reg_T^y$, respectively.
Then, the product distribution of the average strategies $\prn{\frac1T \sumT x_t, \frac1T \sumT y_t}$ is a $(\SReg_T / T)$-approximate Nash equilibrium. 
\end{theorem}

\paragraph{Uncoupled learning dynamics}
In learning in games, each player often employs some form of decentralized algorithm.
The most representative form of this is uncoupled learning dynamics~\citep{hart03uncoupled}:
a learning dynamic is said to be uncoupled if each player's strategy does not depend on the other players' utility functions (\ie~gain or loss functions).
However, in two-player zero-sum games, the $x$-player's utility $\inpr{x, A y}$ and the $y$-player's utility $- \inpr{x, A y}$ differ only in sign, so this condition does not impose any restriction.

\citet{daskalakis11near} introduces the notion of strongly-uncoupled dynamics:
a learning dynamic is said to be strongly-uncoupled if, at each round $t$, the $x$-player determines its strategy using only $(A y_s)_{s=1}^{t-1}$, while the $y$-player determines its strategy using only $(A^\top x_s)_{s=1}^{t-1}$.
Note that under strongly-uncoupled learning dynamics, each player has no access to the opponent's strategies or even to the number of actions available to the opponent.

In this study, we also consider the following intermediate learning dynamics.
A learning dynamic is said to be \emph{cardinality-aware strongly-uncoupled} if, in addition to the information available under strongly-uncoupled dynamics, each player is informed of the number of actions of the opponent.
Note that this definition naturally extends to multiplayer general-sum games.
As we will discuss in \Cref{sec:regret_upper_bounds}, allowing each player to know the opponent's number of actions allows us to improve social and individual regret bounds.

\subsection{Optimistic Hedge}
In the optimistic Hedge algorithm, the strategies of the $x$- and $y$-players are determined as follows:
\begin{equation}
  x_t(i) 
  \propto 
  \exp\prn*{ \eta \prn[\bigg]{ \sum_{s=1}^{t-1} g_s(i) + g_{t-1}(i)} }
  \eqqcolon 
  w_t(i)
  \com
  \
  y_t(i) 
  \propto 
  \exp\prn*{ - \eta' \prn[\bigg]{ \sum_{s=1}^{t-1} \ell_s(i) + \ell_{t-1}(i)} }
  \eqqcolon 
  v_t(i)
  \com
  \label{eq:OptHedge}
\end{equation}
where $\eta, \eta' > 0$ are learning rates of each player, and we set $w_t = v_t = \ones$ and let $g_0 = \ell_0 = \zeros$ for simplicity.
Note that $x_1 = \frac{1}{m} \ones$ and $y_1 = \frac{1}{n} \ones$.
For $t \geq 2$, the update rule can be equivalently written as
\begin{equation}
    x_t(i)
    \propto 
    w_{t-1}(i) \exp\prn*{ \eta \prn*{2 g_{t-1}(i) - g_{t-2}(i)} }
    \com
    \quad
    y_t(i) 
    \propto 
    v_{t-1}(i) \exp\prn*{ - \eta' \prn*{2 \ell_{t-1}(i) - \ell_{t-2}(i)} }
    \per
  \label{eq:OptHedge_onestep}
\end{equation}
The algorithm for the $x$-player is summarized in \Cref{alg:OptHedge}, and the algorithm for the $y$-player can be described analogously.

In \Cref{sec:dynamic}, we provide a theoretical analysis of a learning dynamic based on optimistic Hedge that achieves $O(1/t)$ last-iterate convergence and, as a consequence, an $O(\log T)$ dynamic regret upper bound, slightly improving the bounds in \citet{cai25from}.
Further details of this learning dynamic are given in \Cref{sec:dynamic}.

\LinesNumbered
\SetAlgoVlined  
\begin{algorithm}[t]

\For{$t = 1, 2, \dots, T$}{
  Choose a strategy $x_t \in \Delta_m$ by the optimistic Hedge algorithm in \eqref{eq:OptHedge} or \eqref{eq:OptHedge_onestep}.
  \\
  Observe a gain vector $g_t = A y_t \in [-1, 1]^m$.
}
\caption{
  Optimistic Hedge for the $x$-player
}
\label{alg:OptHedge}
\end{algorithm}

\section{Refining Regret Upper Bounds of Optimistic Hedge}\label{sec:regret_upper_bounds}

This section provides a refined regret analysis of optimistic Hedge to enable a precise comparison with the regret lower bounds derived in the next section.

\subsection{Common Analysis}
We first prepare the following lemma, which generalizes the standard upper bound of optimistic Hedge.
\begin{lemma}\label{lem:regxy_upper}
  Suppose that the $x$- and $y$-players use optimistic Hedge (\Cref{alg:OptHedge}) with learning rates $\eta$ and $\eta'$, respectively.
  Then, for any $c, c' > 0$,
  \begin{align}
    \Reg_T^x
    &\leq
    \frac{\log m}{\eta}
    +
    \frac{\eta}{2 c}  \sumT \nrm{g_t - g_{t-1}}_\infty^2
    -
    \frac{1 - c}{2 \eta}  \sum_{t=2}^T \nrm{x_t - x_{t-1}}_1^2
    \com
    \nn
    \Reg_T^y
    &\leq
    \frac{\log n}{\eta'}
    +
    \frac{\eta'}{2 c'}  \sumT \nrm{\ell_t - \ell_{t-1}}_\infty^2
    -
    \frac{1 - c'}{2 \eta'}  \sum_{t=2}^T \nrm{y_t - y_{t-1}}_1^2
    \per
    \n
  \end{align}
\end{lemma}
The proof is provided in \Cref{app:proof_regret_upper_bounds}.
Here, the parameters $c$ and $c'$ arise from an appropriate decomposition of negative terms that appear in the regret analysis.
Intuitively, the larger these parameters are (within the range up to $1$), the smaller the corresponding player's individual regret becomes.

For $\eta, \eta' > 0$ and $c, c' > 0$, define
\begin{equation}
  \Omega(\eta, \eta', c, c')
  =
  \omega(\eta, c)
  +
  \omega'(\eta', c')
  \com
  \quad
  \omega(\eta, c)
  =
  \frac{\log m}{\eta}
  +
  \frac{\eta}{2 c}
  \com
  \quad
  \omega'(\eta', c')
  =
  \frac{\log n}{\eta'}
  +
  \frac{\eta'}{2 c'}
  \per
  \label{eq:def_omega}
\end{equation}
Then, by \Cref{lem:regxy_upper}, we can derive the following upper bounds on the social regret and the individual regrets.
\begin{theorem}\label{thm:upper_bounds}
Under the assumptions of \Cref{lem:regxy_upper},
for any $c, c' > 0$ satisfying $\eta \eta' \leq \min\set*{c' (1 - c), c (1 - c')}$, 
\begin{align}
  \begin{split}
    \SReg_T
    &\leq
    \Omega(\eta, \eta', c, c')
    \com
    \\
    \Reg_T^x
    &\leq
    \omega(\eta, c)
    +
    \frac{\frac{\eta}{2 c}}{ \frac{1 - c'}{2 \eta'} - \frac{\eta}{2 c} }
    \Omega(\eta, \eta', c, c')
    \eqqcolon
    f(\eta, \eta', c, c')
    \com
    \\
    \Reg_T^y
    &\leq
    \omega'(\eta', c')
    +
    \frac{\frac{\eta'}{2 c'}}{\frac{1 - c}{2 \eta} - \frac{\eta'}{2 c'}}
    \Omega(\eta, \eta', c, c')
    \eqqcolon
    g(\eta, \eta', c, c')
    \per
  \end{split}
  \n
\end{align}
For simplicity, we define $f(\eta, \eta', c, c') = \infty$ when $\eta \eta' = c (1 - c')$, and $g(\eta, \eta', c, c') = \infty$ when $\eta \eta' = c' (1 - c)$.
\end{theorem}
\Cref{thm:upper_bounds} can be proven by the analysis similar to the standard analysis of optimistic Hedge~\citep{rakhlin13optimization,syrgkanis15fast}.
Here, we provide only a proof sketch and defer the complete proof to \Cref{app:proof_regret_upper_bounds}.
\begin{proof}[Proof sketch of \Cref{thm:upper_bounds}]
We first note that 
when $\eta \eta' < c' (1 - c)$, we have
$\frac{\eta'}{2 c'} - \frac{1 - c}{2 \eta} < 0$
and 
when $\eta \eta' < c (1 - c')$ we have 
$\frac{\eta}{2 c} - \frac{1 - c'}{2 \eta'} < 0$.
Hence, summing the two inequalities in \Cref{lem:regxy_upper} and using the inequalities $\nrm{g_t - g_{t-1}}_\infty \leq \nrm{y_t - y_{t-1}}_1$ and $\nrm{\ell_t - \ell_{t-1}}_\infty \leq \nrm{x_t - x_{t-1}}_1$, we have
$
  \SReg_T
  \leq
  \Omega(\eta, \eta', c, c')
  +
  \prn*{\frac{\eta'}{2 c'} - \frac{1 - c}{2 \eta}} 
  \sum_{t=2}^T \nrm{x_t - x_{t-1}}_1^2 
  + 
  \prn*{\frac{\eta}{2 c} - \frac{1 - c'}{2 \eta'}} 
  \sum_{t=2}^T \nrm{y_t - y_{t-1}}_1^2
  \leq
  \Omega(\eta, \eta', c, c')
  ,
$
where the last inequality follows from the assumption that $\eta \eta' \leq \min\set*{c' (1 - c), c (1 - c')}$.
Combining the last upper bound on $\SReg_T$ with the fact that $\SReg_T \geq 0$ gives
$
  \sum_{t=2}^T \nrm{y_t - y_{t-1}}_1^2 
  \leq
  \frac{1}{\frac{1 - c'}{2 \eta'} - \frac{\eta}{2 c}}
  \cdot
  \omega(\eta, \eta', c, c')
  .
$
Here we defined the right-hand side to be $\infty$ whenever $\frac{1 - c'}{2 \eta'} - \frac{\eta}{2 c} = 0$ as in \Cref{thm:upper_bounds}.
Finally, combining \Cref{lem:regxy_upper} with 
$\nrm{g_t - g_{t-1}}_\infty \leq \nrm{y_t - y_{t-1}}_1$
and the last two inequalities, we obtain the desired upper bound on $\Reg_T^x$.
The upper bound on $\Reg_T^y$ can be proven in a similar manner.
\end{proof}
By \Cref{thm:upper_bounds}, the optimal learning rates $\eta, \eta' > 0$ for the social and individual regrets under this analysis can be determined by solving optimization problems  of the functions $\Omega$, $f$, and $g$ over the following feasible region $\Lambda$:
\begin{equation}
  \Lambda
  =
  \set*{
    (\eta, \eta', c, c') \in \Rp^4
    \colon
    \eta \eta' \leq \min\set{c' (1 - c), c (1 - c')}
  }
  \n
\end{equation}
For example, the optimization problem that determines the optimal parameters $(\eta, \eta', c, c') \in \Lambda$ for the social regret is given by $\min_{(\eta, \eta', c, c') \in \Lambda} \Omega(\eta, \eta', c, c')$.
For notational convenience, we sometimes denote an element of $\Lambda$ by $\lambda = (\eta, \eta', c, c')$ and write $M = \log m$ and $N = \log n$ below.
Note that the learning rates $\eta, \eta'$ that minimize the social regret and those that minimize the individual regrets are not necessarily the same.

\subsection{Social Regret Bounds}\label{subsec:social_upper}
We first focus on the social regret.
\begin{lemma}\label{lem:optimal_large_omega}
Let $M' = M + 1/2$, $N' = N + 1/2$, and $D = \sqrt{M' N'} + \sqrt{M N}$.
Then, it holds that
$
  \min_{\lambda \in \Lambda}
  \Omega(\eta, \eta', c, c')
  =
  2 \sqrt{M N'} + 2 \sqrt{M' N}
$.
The minimum is achieved by $\lambda \in \Lambda$ such that
\begin{equation}
  c = c'
  =
  \frac{\sqrt{M' N'}}{D}
  \com \quad
  \eta
  =
  \frac{\sqrt{M M'}}{D}
  \com \quad
  \eta'
  =
  \frac{\sqrt{N N'}}{D}
  \per
  \label{eq:optimal_lr_sreg_casu}
\end{equation}
\end{lemma}
The proof can be found in \Cref{app:proof_regret_upper_bounds}.
Combining \Cref{thm:upper_bounds} with \Cref{lem:optimal_large_omega} yields the following bound:
\begin{theorem}\label{thm:optimal_sreg_casu}
Suppose that the $x$- and $y$-players use optimistic Hedge (\Cref{alg:OptHedge}) with learning rates $\eta$ and $\eta'$ in \Cref{eq:optimal_lr_sreg_casu}. Then,
$
\SReg_T \leq 2 \sqrt{\log m \, (\log n + 1/2)} + 2 \sqrt{\log n \, (\log m + 1/2)}.
$
\end{theorem}
An important remark is that this optimization problem focuses solely on the social regret, and with these choices of learning rates it is not possible to upper bound the individual regrets under the regret analysis via \Cref{thm:upper_bounds}.
This is consistent with the fact that the optimal solution of \Cref{lem:optimal_large_omega} lies on $\eta \eta' = \min\set*{c' (1 - c), c (1 - c')}$, in which case $f = g = \infty$.

In the setting where each player does not know the opponent's number of actions, that is, under strongly-uncoupled learning dynamics, the analysis based on \Cref{thm:upper_bounds} shows that the following social regret cannot be further improved. A rigorous argument is deferred to \Cref{app:proof_regret_upper_bounds}.
\begin{theorem}\label{thm:optimal_sreg_su}
Suppose that the $x$- and $y$-players use optimistic Hedge (\Cref{alg:OptHedge}) with learning rates $\eta = \eta' = 1/2$. Then,
$
\SReg_T \leq 2 \log (m n) + 1
$.
\end{theorem}
Note that \Cref{thm:optimal_sreg_su} is not a new result, although no prior literature has explicitly stated this upper bound with the leading constants and investigated optimal leading constants under this analysis.
Our cardinality-aware upper bound in \Cref{thm:optimal_sreg_casu} is strictly better than the one in \Cref{thm:optimal_sreg_su}.
In fact, by the AM--GM inequality, we have
$
2 \sqrt{\log m \, (\log n + 1/2)} + 2 \sqrt{\log n \, (\log m + 1/2)}
\leq
2 \log (m n) + 1
$.
As suggested by the nature of the AM--GM inequality, this implies that the advantage of being cardinality-aware becomes more significant as $\max\set{\log m / \log n, \log n / \log m}$ increases, which is particularly illustrated in the examples mentioned in \Cref{sec:introduction}.

\begin{wrapfigure}[12]{r}{.52\textwidth}
  \vspace{-\baselineskip}
  \centering
  \begin{minipage}[t]{.49\linewidth}
    \centering
    \includegraphics[width=\linewidth]{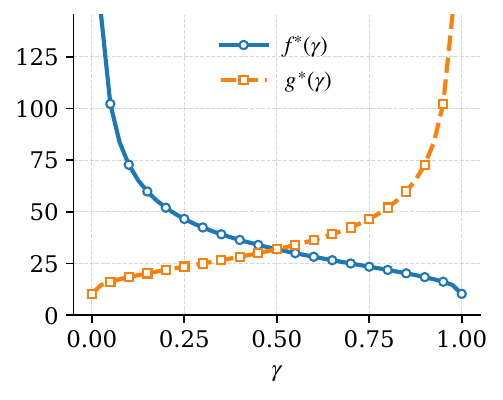}
  \end{minipage}\hfill
  \begin{minipage}[t]{.49\linewidth}
    \centering
    \includegraphics[width=\linewidth]{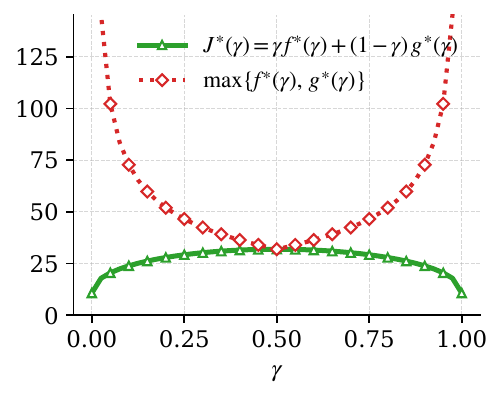}
  \end{minipage}
  \vspace{-5pt}
  \caption{
    Tradeoff versus $\gamma\in(0,1)$ when $m = n = 10^2$: 
    (a) $f^*(\gamma) = f(\lambda_\gamma)$ and $g^*(\gamma) = g(\lambda_\gamma)$ for
    $\lambda_{\gamma} = \argmin_{\lambda\in\Lambda}J_{\gamma}(\lambda)$;
    (b) $J^{\ast}(\gamma) = J_\gamma(\lambda_\gamma)$ and $\max\{f^{\ast},g^{\ast}\}$.
  }
  \label{fig:indiv_reg_vs_gamma}
  \vspace{-13pt}
\end{wrapfigure}

\subsection{Individual Regret Bounds}\label{subsec:indiv_upper}
Next, we turn our focus to the individual regret.
Note that while $\Reg_T^x \leq \min_{\lambda \in \Lambda} f(\lambda)$ and $\Reg_T^y \leq \min_{\lambda \in \Lambda} g(\lambda)$ hold, the learning rates that achieve the minimum on the right-hand sides of the two inequalities are not necessarily the same.
Thus, we need to determine the criterion by which $\lambda$ (and thus learning rates $\eta, \eta'$) is chosen.

A natural objective is to minimize the maximum of the individual regrets, $\max \set{\Reg_T^x, \Reg_T^y}$ by solving,
$\min_{\lambda \in \Lambda} \max \set{f(\lambda), g(\lambda)}$.
One may also be interested in how much one player's regret can be minimized at the expense of the other player's regret, which is useful for comparison with the lower bounds derived later.
To treat these cases in a unified manner, we consider minimizing a weighted sum of $f$ and $g$: for $\gamma \in [0,1]$,
\begin{equation}
  J_\gamma(\lambda)
  =
  \gamma \,
  f(\lambda)
  +
  (1 - \gamma) \,
  g(\lambda)
  \per
  \label{eq:def_Jgamma}
\end{equation}
The optimization problem for $J_\gamma$ is nonconvex with respect to $\lambda$, and unlike the case of the social regret analysis, it does not admit a closed-form solution.
However, by introducing an appropriate change of variables and applying gradient-based methods to these variables, we can transform this nonconvex optimization problem to a convex optimization problem,
and thus it is possible to numerically compute the optimal values and solutions of $f$ and $g$ for each $\gamma \in [0,1]$.
Therefore, even when a closed-form solution is not available, solving this convex problem enables us to obtain desirable learning rates $\eta$ and $\eta'$.
The resulting optimal values of $f$ and $g$ for each $\gamma$ computed by solving the convex optimization problem are shown in \Cref{fig:indiv_reg_vs_gamma}.
This figure illustrates the tradeoff between the individual regrets of the $x$- and $y$-players.


\paragraph{Extreme cases}
Here we discuss only the upper bound in the extreme case of this tradeoff.
If we optimize solely for $f$, that is, if we choose $\eta, \eta', c, c'$ to minimize only the $x$-player's regret, then the parameters approach $c \to 1$, $c' \to 0$, $\eta = \sqrt{M / (N + 1/2)}$, and $\eta' \to 0$.
In this case, the regret of the $x$-player is bounded by
\begin{equation}
  \Reg_T^x
  \leq
  f(\lambda)
  \to
  \frac{M}{\eta}
  +
  \frac{\eta}{2}
  +
  \eta N
  =
  2 \sqrt{M (N + 1/2)}
  \per
  \label{eq:x_best_indiv_casu}
\end{equation}
As we will see in \Cref{sec:lower_bound}, this result exhibits exactly a factor of $2$ gap compared with the corresponding lower bound.

In the case without cardinality-awareness, replacing the above $\eta$ with $\eta = 1$ yields
\begin{equation}
  \Reg_T^x
  \leq
  f(\lambda)
  \to
  \frac{M}{\eta}
  +
  \frac{\eta}{2}
  +
  \eta N
  =
  M + N + \frac12
  \per
  \label{eq:x_best_indiv_su}
\end{equation}
As we will see in \Cref{sec:lower_bound}, this corresponds to an additive $\log n$ factor gap compared with the corresponding lower bound.

These parameter settings correspond to the following learning dynamic: 
the $x$-player runs optimistic Hedge with a learning rate of either $\eta = \sqrt{M / (N + 1/2)}$ or $\eta = 1$, while the $y$-player runs optimistic Hedge with a learning rate of zero (equivalent to playing the uniform strategy at every round).
Although \Cref{thm:upper_bounds} cannot be applied directly to analyze the algorithm in this limiting case, the upper bounds in \Cref{eq:x_best_indiv_casu,eq:x_best_indiv_su} can be obtained directly from \Cref{lem:regxy_upper} (the proof can be found in \Cref{app:proof_regret_upper_bounds}).

\paragraph{Bounding max of individual regrets}
It is also important to upper bound the maximum of the two players' individual regrets.
A closed-form expression is not available in general, and thus we derive an upper bound on the optimal value focusing on the case of $\gamma = 1/2$, thereby upper bounding the maximum individual regret.
By applying an appropriate change of variables, we can prove the following bounds.
\begin{lemma}\label{lem:optimal_J_gamma}
Let $M' = M + 1/2$, $N' = N + 1/2$, and $D = \sqrt{M' N'} + \sqrt{M N}$.
Then, 
$
\min_{\lambda \in \Lambda}
\max\set{f(\lambda), g(\lambda)}
\leq
2
\min_{\lambda \in \Lambda} J_{1/2}(\lambda)
\leq
(20 / 3)
\prn{
  \sqrt{M N'}
  +
  \sqrt{M' N}
}
$,
where $J_{1/2}(\eta, \eta', c, c') \leq \text{(RHS)}$ holds when
\begin{equation}
  \eta
  =
  \frac{\sqrt{M M'}}{2 D}
  \com
  \quad
  \eta'
  =
  \frac{\sqrt{N N'}}{2 D}
  \per
  \label{eq:optimal_lr_maxreg_casu}
\end{equation}
\end{lemma}
The proof can be found in \Cref{app:proof_regret_upper_bounds}.
Combining \Cref{thm:upper_bounds} with \Cref{lem:optimal_J_gamma} immediately yields the following result:
\begin{theorem}\label{thm:optimal_max_indivreg_casu}
Suppose that the $x$- and $y$-players use optimistic Hedge (\Cref{alg:OptHedge}) with learning rates $\eta$ and $\eta'$ in \Cref{eq:optimal_lr_maxreg_casu}. Then,
$
\max\set{\Reg_T^x, \Reg_T^y}
\leq
(20/3) \prn{ \sqrt{\log m (\log n + 1/2)} + \sqrt{\log n (\log m + 1/2)} }
.
$
\end{theorem}
In the above analysis, an unnecessary gap arises in the first inequality of \Cref{lem:optimal_J_gamma}. By directly bounding $\max\set{f(\lambda), g(\lambda)}$ instead of bounding the minimizer of $J_{1/2}$, one can obtain some improvement in leading constants. 

As in the case of social regret, under the (cardinality-unaware) strongly-uncoupled learning dynamics, the analysis based on \Cref{thm:upper_bounds} cannot yield a bound better than the following maximum of the individual regret bound. 
\begin{theorem}\label{thm:optimal_max_indivreg_su}
Suppose that the $x$- and $y$-players use optimistic Hedge (\Cref{alg:OptHedge}) with learning rates $\eta = \eta' = 1/(2 \sqrt{3})$ (and $c = c' = 1/2$). Then,
$
\max\set{\Reg_T^x, \Reg_T^y} \leq 3 \sqrt{3} \log (m n) + {1}/{\sqrt{3}}.
$
\end{theorem}
A rigorous argument is provided in \Cref{app:proof_regret_upper_bounds}.
Note that in the cardinality-unaware case, for both the analysis of the social regret in \Cref{thm:optimal_sreg_su} and the analysis of the maximum of the individual regrets in \Cref{thm:optimal_max_indivreg_su}, the corresponding optimization problems $\min_\lambda \Omega(\lambda)$ and $\min_\lambda \max\set{f(\lambda), g(\lambda)}$ admit closed-form optimal solutions.
This implies that, in the cardinality-unaware setting, the leading constants of the regret upper bounds obtained in \Cref{thm:optimal_sreg_su,thm:optimal_max_indivreg_su} cannot be further improved as long as one relies on the commonly used regret analysis approach~\citep{rakhlin13optimization,syrgkanis15fast,anagnostides22near,anagnostides22uncoupled}.
As we will see below, this limitation is not an issue for the social regret, since the upper and lower bounds match including leading constants; however, for the individual regret, there is room to sharpen both the upper or the lower bounds.
\section{Regret Lower Bounds}\label{sec:lower_bound}
This section investigates individual regret lower bounds when each player uses optimistic Hedge in learning in two-player zero-sum games.
Our main result is as follows.
\begin{theorem}\label{thm:lower_bounds}
  Suppose that the $x$- and $y$-players use the optimistic Hedge algorithm (\Cref{alg:OptHedge}) with learning rates $\eta, \eta' > 0$.
  Then, there exists an instance of learning in two-player zero-sum games such that 
  the regret of each player is lower bounded as
  \begin{equation}
    \Reg^x_T 
    \geq
    \begin{cases}
      \displaystyle
      \frac{\log m}{\eta}
      -
      \frac{\log ((m - 1)(T + 1)) + 1}{\eta(T + 1)} 
      & 
      \displaystyle
      \mbox{if} \ \
      \eta \geq \frac{\log ((m - 1)(T + 1))}{(T + 1)}
      \com
      \vspace{7pt}
      \\ 
      \displaystyle
      \frac{1}{\eta}
      \prn*{
        \log m 
        -
        \eta
        -
        (m - 1) \e^{- \eta (T+1)}
      }      
      &
      \displaystyle
      \mbox{if} \ \
      \eta \in \bigg(0, \frac{\log ((m - 1)(T + 1))}{(T + 1)} \bigg]
      \com
    \end{cases}
    \n
  \end{equation}
  \begin{equation}
    \Reg^y_T 
    \geq
    \begin{cases}
      \displaystyle
      \frac{\log n}{\eta'}
      -
      \frac{\log ((n - 1)(T + 1)) + 1}{\eta(T + 1)} 
      \displaystyle
      & 
      \displaystyle
      \mbox{if} \ \
      \eta' \geq \frac{\log ((n - 1)(T + 1))}{(T + 1)}
      \com
      \vspace{7pt}
      \\ 
      \displaystyle
      \frac{1}{\eta'}
      \prn*{
        \log n
        -
        \eta'
        -
        (n - 1) \e^{- \eta' (T+1)}
      }      
      & 
      \displaystyle
      \mbox{if} \ \
      \eta' \in \bigg(0, \frac{\log ((n - 1)(T + 1))}{(T + 1)} \bigg]
      \per
    \end{cases}
    \n
  \end{equation}
\end{theorem}
To our knowledge, this work is the first to derive regret lower bounds for optimistic Hedge and to investigate their dependence on the numbers of actions $m$ and $n$.
Note that the social regret lower bound can be obtained directly from the individual regret lower bounds.

We compare the lower bounds in \Cref{thm:lower_bounds} with the regret upper bounds in \Cref{sec:regret_upper_bounds}.
We begin with the case of cardinality-aware strongly-uncoupled learning dynamics (summarized in the lower half of \Cref{table:regret}).
For simplicity, we focus only on the leading terms.
In this case, the social regret upper bound is $2 \sqrt{\log m \log n}$ for $\eta, \eta'$ in \Cref{eq:optimal_lr_sreg_casu}, which matches the lower bound $2 \sqrt{\log m \log n} - o(1)$ from \Cref{thm:lower_bounds}, including the leading constant.
On the other hand, for the individual regret upper bound, even if we focus solely on minimizing regret of the $x$-player, with $\eta, \eta'$ chosen as in \Cref{eq:x_best_indiv_casu}, the individual regret asymptotically becomes $2 \sqrt{\log m \log n}$, which exhibits exactly a factor-of-two gap compared with the lower bound $\sqrt{\log m \log n} - o(1)$ obtained from \Cref{thm:lower_bounds}.

Next, we consider the case of strongly-uncoupled learning dynamics (summarized in the upper half of \Cref{table:regret}).
In this case, the social regret upper bound is $2 \log (m n)$ for $\eta = \eta' = 1/2$, which matches the lower bound $2 \log (m n) - o(1)$ from \Cref{thm:lower_bounds}, including the leading constant.
On the other hand, for the individual regret upper bound, even with $\eta, \eta'$ chosen as in \Cref{eq:x_best_indiv_su}, it can only become $\log (m n)$ asymptotically.
This leaves an additive gap of $\log n$ compared with $\log m - o(1)$ from \Cref{thm:lower_bounds}.
Closing the gaps between the upper and lower bounds for individual regret remains an important direction for future work.

\subsection{Proof of \Cref{thm:lower_bounds}}
Here we provide the proof of \Cref{thm:lower_bounds}.
We use the following inequality to prove the theorem:
\begin{lemma}\label{lem:ineq_telescope}
  For any $z > 0$ and $a > 0$,
  \begin{equation}
    \frac{z}{1+z}
    \geq
    \frac{1}{a}
    \brk*{
      \log (1 + z)
      -
      \log (1 + z \e^{-a})
    }
    \n
    \per
  \end{equation}
\end{lemma}
The proof is provided in \Cref{app:proof_lower_bound}.
Now, we are ready to prove \Cref{thm:lower_bounds}.
\begin{proof}[Proof of \Cref{thm:lower_bounds}]
We consider a game with the following payoff matrix $A \in [-1, 1]^{m \times n}$:
\begin{equation}
  A(i,j)
  =
  \begin{cases}
    0 & \mbox{if} \, (i,j) = (1,1) \com \\
    \Delta & \mbox{if} \, i = 1, j \neq 1 \com \\
    - \Delta & \mbox{if} \, i \neq 1, j = 1 \com \\
    0 & \mbox{otherwise} \com
  \end{cases}
  \label{eq:def_A}
\end{equation}
for $\Delta \in (0, 1]$.
Then, we have
\begin{equation}
  \begin{split}    
    g_t 
    &= 
    A y_t
    =
    \prn{ \Delta(1 - y_t(1)), - \Delta y_t(1), \dots, - \Delta y_t(1)}^\top
    \com
    \nn
    \ell_t 
    &= 
    A^\top x_t
    =
    \prn{ - \Delta(1 - x_t(1)), \Delta x_t(1), \dots, \Delta x_t(1)}^\top
    \per
    \n
  \end{split}
\end{equation}
We also have
$
\inpr{x_t, A y_t}
=
\Delta \prn{x_t(1) - y_t(1)}.
$
Hence, 
since actions $1$ are optimal for both players,
the regret of the $x$-player can be rewritten as
\begin{equation}
  \Reg^x_T
  =
  \sumT \Delta (1 - x_t(1))
  \per
  \label{eq:reg_for_A}
\end{equation}

In what follows, we will evaluate $x_t(1)$ and $y_t(1)$.
Fix arbitrary $k \in [m] \setminus \set{1}$.
Then, recalling that the update rule of the optimistic Hedge algorithm can be written as \Cref{eq:OptHedge_onestep},
for any $t \geq 3$ we have
\begin{equation}
  \frac{w_t(k)}{w_t(1)}
  = 
  \frac{w_{t - 1}(k)}{w_{t - 1}(1)}  
  \exp\brk*{
    \eta \prn*{
      2 \prn{ g_{t - 1}(k) - g_{t - 1}(1) }
      -
      \prn{ g_{t - 2}(k) - g_{t - 2}(1) }
    }
  }
  = %
  \frac{w_{t - 1}(k)}{w_{t - 1}(1)}  
  \exp\prn*{
    - \eta \Delta
  }
  \per
  \n
\end{equation}
Repeatedly applying the last equality and noting that $g_0 = \zeros$, we have
\begin{equation}
  \frac{w_t(k)}{w_t(1)}
  =
  \frac{w_2(k)}{w_2(1)}  
  \exp\prn*{
    - \eta \Delta (t - 2) 
  }
  =
  \frac{w_1(k)}{w_1(1)}  
  \exp\prn*{
    - \eta \Delta t
  }
  =
  \exp\prn*{
    - \eta \Delta t
  }
  \com
\end{equation}
where we used $w_1(i) = 1$ for all $i \in [m]$.
From this equality, we have
\begin{align}
  \frac{1}{x_t(1)}
  =
  \frac{w_t(1) + \sum_{i \in [m] \setminus \set{1}} w_t(i)}{w_t(1)}
  =
  1
  +
  (m-1)
  \e^{
    - \eta \Delta t
  }
  \com
  \n
\end{align}
which implies that for each $t \in [t]$ it holds that
\begin{equation}
  x_t(1)
  =
  \prn{1 + \alpha_t}^{-1}
  \com
  \ \
  \alpha_t
  \coloneqq
  (m-1)
  \exp\prn*{
    - \eta \Delta t
  }
  \per
  \label{eq:xtone_for_A}
\end{equation}
Finally, combining \Cref{eq:reg_for_A} with \Cref{eq:xtone_for_A},
we can lower bound the regret of the $x$-player as
\begin{align}
  \Reg^x_T
  &\geq
  \sumT \Delta \prn*{1 - \frac{1}{1 + \alpha_t}}
  =
  \Delta
  \sumT \frac{\alpha_t}{1 + \alpha_t}
  \nn
  &\geq
  \Delta \sumT 
  \frac{1}{\eta \Delta}
  \prn*{
    \log (1 + \alpha_t)
    -
    \log (1 + \alpha_{t+1})
  }
  \tag{\Cref{lem:ineq_telescope} with $z = \alpha_t$ and $a = \eta \Delta$}
  \nn
  &=
  \frac{1}{\eta}
  \prn*{
    \log (1 + \alpha_1)
    -
    \log (1 + \alpha_{T+1})
  }
  \nn
  &\geq
  \frac{1}{\eta}
  \prn*{
    \log m 
    -
    \eta \Delta
    -
    (m - 1) \e^{- \eta \Delta (T+1)}
  }
  \com
  \label{eq:reg_x_lower_proof_1}
\end{align}
where in the last inequality we used
$\log(1 + \alpha_1) = \log (1 + (m - 1) \exp(- \eta \Delta)) \geq \log (m \exp(- \eta \Delta))$
and $\log(1 + z) \leq z$ for $z \in \R$.
Using the fact that the function $g \colon (0, 1] \to \R$ given by
$
  g(\Delta)
  =
  \eta \Delta
  +
  (m - 1) \exp(- \eta \Delta (T+1))
$
is minimized when 
$
  \Delta^*
  =
  \min\set*{
    1,
    \frac{\log \prn{(m - 1)(T + 1)}}{\eta (T + 1)}
  }
$
and its optimal value $g(\Delta^*)$ is 
\begin{equation}
  \begin{cases}
    \displaystyle
    \frac{\log ((m - 1)(T + 1))}{T + 1} 
    & 
    \displaystyle
    \mbox{if} \ \eta \geq \frac{\log ((m - 1)(T + 1))}{T + 1} 
    \com
    \\ 
    \displaystyle
    \eta + (m - 1) \e^{- \eta (T + 1)} 
    & 
    \mbox{otherwise}
    \com
  \end{cases}
  \n
\end{equation}
choosing $\Delta = \Delta^*$ in \Cref{eq:reg_x_lower_proof_1} gives the desired lower bound for the $x$-player.
The regret of the $y$-player can be lower bounded by the same argument.
\end{proof}

\section{Dynamic Regret Bounds}\label{sec:dynamic}
This section provides upper and lower bounds on dynamic regret, which are closely related to last-iterate convergence.

\subsection{Dynamic Regret Upper Bounds}
We describe a learning dynamic that guarantees $\tilde{O}(1/T)$ last-iterate convergence by outputting the average of the optimistic Hedge iterates~\citep{cai25from}, which implies that each player's dynamic regret can be bounded by $\tilde{O}(\log T)$.
In this learning dynamic, each player first uses optimistic Hedge to compute $\xhat_t \in \Delta_m$ and $\yhat_t \in \Delta_n$ as follows:
\begin{equation}
  \begin{split}    
    \xhat_t(i) 
    &\propto
    \exp\prn*{ \eta \prn[\bigg]{ \sum_{s=1}^{t-1} \ghat_s(i) + \ghat_{t-1}(i)  }  }
    \com \quad
    \ghat_t = A \xhat_t 
    \com
    \\
    \yhat_t(i) 
    &\propto
    \exp\prn*{  - \eta' \prn[\bigg]{ \sum_{s=1}^{t-1} \lhat_s(i) + \lhat_{t-1}(i)  }  }
    \com \quad
    \lhat_t = A^\top \yhat_t 
    \per
  \end{split}
  \label{eq:def_xhat_yhat}
\end{equation}
Then, each player adopts the average of these past outputs as their strategies:
\begin{equation}
  x_t = \frac{1}{t} \sum_{s=1}^t \xhat_s
  \com
  \quad
  y_t = \frac{1}{t} \sum_{s=1}^t \yhat_s
  \per
  \label{eq:def_average_iter}
\end{equation}
Their key observation is that, using the gradients defined by the actual average strategies $x_t, y_t$, namely $g_t = A x_t$ and $\ell_t = A^\top y_t$, one can reconstruct the gradients $\ghat_t = A \xhat_t$ and $\lhat_t = A^\top \yhat_t$ (\ie~the gradients that would have been obtained if the optimistic Hedge outputs $\xhat_t, \yhat_t$ had been used directly as their strategies) as follows:
\begin{equation}
  \ghat_t = t \cdot g_t - \sum_{s=1}^{t-1} \ghat_s
  \com\quad
  \lhat_t = t \cdot \ell_t - \sum_{s=1}^{t-1} \lhat_s
  \per
  \label{eq:utility_estimate}
\end{equation}
In fact, the quantities $\ghat_t$ and $\lhat_t$ can be computed from the information available up to time $t-1$ together with the gradients of the average strategies $g_t = A x_t$ and $\ell_t = A^\top y_t$.
By induction and using \Cref{eq:def_xhat_yhat,eq:def_average_iter}, we obtain
$
t \cdot g_t - \sum_{s=1}^{t-1} \ghat_s
=
t \cdot A \prn[\big]{\frac{1}{t} \sum_{s=1}^t \xhat_s} - \sum_{s=1}^{t-1} \ghat_s
=
\ghat_t
$
which verifies the equality in \Cref{eq:utility_estimate}.
The algorithm for the $x$-player is summarized in \Cref{alg:OptHedge_lastiterate}. 

\LinesNumbered
\SetAlgoVlined  
\begin{algorithm}[t]

\For{$t = 1, 2, \dots, T$}{
  Compute a strategy $\xhat_t \in \Delta_m$ by the optimistic Hedge algorithm in \eqref{eq:def_xhat_yhat}.
  \\
  Choose the time-averaged strategy $x_t$ in \eqref{eq:def_average_iter}.
  \\
  Observe a gain vector $g_t = A y_t \in [-1, 1]^m$, where $y_t$ is given by \eqref{eq:def_average_iter}.
  \\
  Recover $\hat{g}_t \in [-1, 1]^m$ by \eqref{eq:utility_estimate}.
}
\caption{
  Algorithm based on optimistic Hedge for the $x$-player with $\tilde{O}(\log T)$ dynamic regret
}
\label{alg:OptHedge_lastiterate}
\end{algorithm}

From the above observation and \Cref{thm:sreg_to_nash}, we see that this dynamic achieves the following last-iterate convergence and dynamic regret bound:
\begin{theorem}[{\citealt[Theorem 3]{cai25from}}]\label{thm:cai_lastiterate}
  Let $(x_t)_t$ and $(y_t)_t$ be sequences of strategies generated by \Cref{alg:OptHedge_lastiterate} with $\eta = \eta' = 1/2$.
  Then for any $t \geq 1$, the product distribution $(x_t, y_t)$ is an $(2 \log(mn) / t)$-approximate Nash equilibrium.
  Consequently, the dynamic regret of each player is upper bounded as
  \\ $\max\set{\DReg_T^x, \DReg_T^y} \leq (2 \log (m n) + 1) (\log T + 1)$.
\end{theorem}
Under cardinality-aware strongly-uncoupled learning dynamics, the bounds of \Cref{thm:cai_lastiterate} can be improved as follows by using the improve social regret bound in \Cref{thm:optimal_sreg_casu}:
\begin{theorem}\label{thm:dynamic_reg_casu}
  Let $(x_t)_t$ and $(y_t)_t$ be sequences of strategies generated by \Cref{alg:OptHedge_lastiterate} with $\eta, \eta'$ in \Cref{eq:optimal_lr_sreg_casu}.
  Then for any $t \geq 1$, the product distribution $(x_t, y_t)$ is an $(2 \sqrt{\log m \, (\log n + 4)} + 2 \sqrt{\log n \, (\log m + 4)} / t)$-approximate Nash equilibrium.
  Consequently, the dynamic regrets are bounded as
  $\max\set{\DReg_T^x, \DReg_T^y} 
  \leq 
  2 \prn{\sqrt{\log m \, (\log n + 1/2)} + \sqrt{\log n \, (\log m + 1/2)}} (\log T + 1)$.
\end{theorem}
Note that, by the AM--GM inequality, the convergence rate in \Cref{thm:dynamic_reg_casu} is better than that in \Cref{thm:cai_lastiterate}.

\subsection{Dynamic Regret Lower Bounds}
Here we provide dynamic regret lower bounds for the learning dynamic based on optimistic Hedge in \Cref{alg:OptHedge_lastiterate}.
\begin{theorem}\label{thm:lower_bounds_dynamic}
  Suppose that the $x$- and $y$-players use \Cref{alg:OptHedge_lastiterate} with learning rates $\eta, \eta' > 0$.
  Let $\kappa(T) = \sqrt{T + 1} + 1$.
  Then, there exists an instance of learning in two-player zero-sum games such that 
  the dynamic regret of each player is lower bounded as
  \begin{equation}
    \DReg^x_T 
    \geq
    \begin{cases}
      \displaystyle
      \frac{\log m \log (T + 1)}{2 \eta}
      -
      \frac{\log ((m - 1) \kappa(T)) + 1}{\eta \, \kappa(T)} 
      & \displaystyle \mbox{if} \ \
      \eta \geq \frac{\log ((m - 1) \kappa(T))}{\kappa(T)}
      \com
      \vspace{7pt}
      \\ 
      \displaystyle
      \frac{\log (T + 1)}{2 \eta}
      \prn*{
        \log m 
        -
        \eta
        -
        (m - 1) \e^{- \eta \kappa(T)}
      }      
      & \displaystyle \mbox{if} \ \
      \eta \in \bigg(0, \frac{\log ((m - 1) \kappa(T) )}{\kappa(T)} \bigg]
      \com
    \end{cases}
    \n
  \end{equation}
  \begin{equation}
    \DReg^y_T 
    \geq
    \begin{cases}
      \displaystyle
      \frac{\log n \log (T + 1)}{2 \eta'}
      -
      \frac{\log ((n - 1) \kappa(T)) + 1}{\eta' \kappa(T)} 
      & \displaystyle \mbox{if} \ \
      \eta' \geq \frac{\log ((n - 1) \kappa(T))}{\kappa(T)}
      \com
      \vspace{7pt}
      \\ 
      \displaystyle
      \frac{\log (T + 1)}{2 \eta'}
      \prn*{
        \log n
        -
        \eta'
        -
        (n - 1) \e^{- \eta' \kappa(T)}
      }      
      & \displaystyle \mbox{if} \ \
      \eta' \in \bigg(0, \frac{\log ((n - 1)\, \kappa(T))}{\kappa(T)} \bigg]
      \per
    \end{cases}
    \n
  \end{equation}
\end{theorem}
The proof is provided in \Cref{app:proof_dynamic}.
A comparison with the dynamic regret upper bounds is summarized in \Cref{table:regret}.
These results show that, for the learning dynamics presented above, the bounds are nearly optimal with respect to the numbers of actions $m$, $n$, and the number of rounds $T$.
In the analysis of the dynamic regret lower bound, it is necessary to extract not only the logarithmic factor in the number of actions but also an additional $\log T$ factor, which worsens the constant in the lower bound by a factor of two compared with that of the external regret lower bound in \Cref{thm:lower_bounds}.

\section{Conclusion and Future Work}\label{sec:conclusion}
In this paper, we investigated the regret upper and lower bounds of learning dynamics based on optimistic Hedge, one of the most representative dynamics for learning in two-player zero-sum games.
Specifically, we first refined the regret upper bounds of optimistic Hedge.
We then derived algorithm-dependent regret lower bounds, showing that most of these upper bounds are in fact optimal, and that the social regret is optimal even with respect to the leading constant.
Finally, we extended these techniques to provide an improved upper bound and a new lower bound for dynamic regret.

This paper opens several interesting directions for future research.
The first is to investigate the intermediate regimes between uncoupled learning dynamics and strongly-uncoupled learning dynamics in multiplayer general-sum games, such as the proposed cardinality-aware strongly-uncoupled learning dynamics.
We have shown that allowing players to know the opponent's number of actions leads to improved regret bounds.
It is an interesting question whether such improvements also extend to external regret minimization~\citep{anagnostides22near} and swap regret minimization~\citep{anagnostides22uncoupled,tsuchiya25corrupted} in multiplayer general-sum games.
Moreover, in our analysis of the individual and dynamic regret, the upper and lower bounds still exhibit a certain gap, and investigating whether this gap can be closed remains an important direction for future work.

A more important direction for future work is to investigate the dependence on the numbers of actions $m$ and $n$ for general strongly-uncoupled learning dynamics in two-player zero-sum games.
A limitation of this paper is that the derived regret lower bounds are specific to the learning dynamics based on optimistic Hedge.
One possible direction is to derive lower bounds separately for dynamics that possess a certain form of stability and for those that do not.
For example, follow-the-leader (or fictitious play), which is an unstable algorithm, achieves $O(1)$ regret on the payoff matrix used in our lower-bound construction.
However, as one might naturally expect, follow-the-leader suffers linear regret against an appropriately constructed payoff matrix.
Accordingly, it may be a fruitful approach to distinguish between algorithms that exhibit a certain stability property (such as Hedge) and those that lack such stability, and to analyze them separately.

\bibliography{references.bib}

\begin{thebibliography}{35}
\providecommand{\natexlab}[1]{#1}
\providecommand{\url}[1]{\texttt{#1}}
\expandafter\ifx\csname urlstyle\endcsname\relax
  \providecommand{\doi}[1]{doi: #1}\else
  \providecommand{\doi}{doi: \begingroup \urlstyle{rm}\Url}\fi

\bibitem[Anagnostides et~al.(2022{\natexlab{a}})Anagnostides, Daskalakis, Farina, Fishelson, Golowich, and Sandholm]{anagnostides22near}
Ioannis Anagnostides, Constantinos Daskalakis, Gabriele Farina, Maxwell Fishelson, Noah Golowich, and Tuomas Sandholm.
\newblock Near-optimal no-regret learning for correlated equilibria in multi-player general-sum games.
\newblock In \emph{Proceedings of the 54th Annual ACM SIGACT Symposium on Theory of Computing}, page 736^^e2^^80^^93749. Association for Computing Machinery, 2022{\natexlab{a}}.

\bibitem[Anagnostides et~al.(2022{\natexlab{b}})Anagnostides, Farina, Kroer, Lee, Luo, and Sandholm]{anagnostides22uncoupled}
Ioannis Anagnostides, Gabriele Farina, Christian Kroer, Chung-Wei Lee, Haipeng Luo, and Tuomas Sandholm.
\newblock Uncoupled learning dynamics with {$O(\log T)$} swap regret in multiplayer games.
\newblock In \emph{Advances in Neural Information Processing Systems}, volume~35, pages 3292--3304. Curran Associates, Inc., 2022{\natexlab{b}}.

\bibitem[Bowling et~al.(2015)Bowling, Burch, Johanson, and Tammelin]{bowling15heads}
Michael Bowling, Neil Burch, Michael Johanson, and Oskari Tammelin.
\newblock Heads-up limit hold'em poker is solved.
\newblock \emph{Science}, 347\penalty0 (6218):\penalty0 145--149, 2015.

\bibitem[Boyd and Vandenberghe(2004)]{boyd04convex}
Stephen~P Boyd and Lieven Vandenberghe.
\newblock \emph{Convex optimization}.
\newblock Cambridge university press, 2004.

\bibitem[Cai et~al.(2025)Cai, Luo, Wei, and Zheng]{cai25from}
Yang Cai, Haipeng Luo, Chen-Yu Wei, and Weiqiang Zheng.
\newblock From average-iterate to last-iterate convergence in games: A reduction and its applications.
\newblock In \emph{Advances in Neural Information Processing Systems}, volume~29, 2025.

\bibitem[Cesa-Bianchi and Lugosi(2006)]{cesabianchi06prediction}
Nicolo Cesa-Bianchi and G{\'a}bor Lugosi.
\newblock \emph{Prediction, learning, and games}.
\newblock Cambridge university press, 2006.

\bibitem[Chen and Peng(2020)]{chen20hedging}
Xi~Chen and Binghui Peng.
\newblock Hedging in games: Faster convergence of external and swap regrets.
\newblock In \emph{Advances in Neural Information Processing Systems}, volume~33, pages 18990--18999. Curran Associates, Inc., 2020.

\bibitem[Daskalakis and Panageas(2019)]{daskalakis19last}
Constantinos Daskalakis and Ioannis Panageas.
\newblock Last-iterate convergence: Zero-sum games and constrained min-max optimization.
\newblock In \emph{10th Innovations in Theoretical Computer Science conference}, 2019.

\bibitem[Daskalakis et~al.(2011)Daskalakis, Deckelbaum, and Kim]{daskalakis11near}
Constantinos Daskalakis, Alan Deckelbaum, and Anthony Kim.
\newblock Near-optimal no-regret algorithms for zero-sum games.
\newblock In \emph{Proceedings of the Twenty-Second Annual ACM-SIAM Symposium on Discrete Algorithms}, page 235^^e2^^80^^93254. Society for Industrial and Applied Mathematics, 2011.

\bibitem[{FAIR} et~al.(2022){FAIR}, Bakhtin, Brown, Dinan, Farina, Flaherty, Fried, Goff, Gray, Hu, Jacob, Komeili, Konath, Kwon, Lerer, Lewis, Miller, Mitts, Renduchintala, Roller, Rowe, Shi, Spisak, Wei, Wu, Zhang, and Zijlstra]{meta22human}
Meta Fundamental AI Research Diplomacy~Team {FAIR}, Anton Bakhtin, Noam Brown, Emily Dinan, Gabriele Farina, Colin Flaherty, Daniel Fried, Andrew Goff, Jonathan Gray, Hengyuan Hu, Athul~Paul Jacob, Mojtaba Komeili, Karthik Konath, Minae Kwon, Adam Lerer, Mike Lewis, Alexander~H. Miller, Sasha Mitts, Adithya Renduchintala, Stephen Roller, Dirk Rowe, Weiyan Shi, Joe Spisak, Alexander Wei, David Wu, Hugh Zhang, and Markus Zijlstra.
\newblock Human-level play in the game of {{D}}iplomacy by combining language models with strategic reasoning.
\newblock \emph{Science}, 378\penalty0 (6624):\penalty0 1067--1074, 2022.

\bibitem[Foster et~al.(2016)Foster, Li, Lykouris, Sridharan, and Tardos]{foster16learning}
Dylan~J Foster, Zhiyuan Li, Thodoris Lykouris, Karthik Sridharan, and Eva Tardos.
\newblock Learning in games: Robustness of fast convergence.
\newblock In \emph{Advances in Neural Information Processing Systems}, volume~29, pages 4734--4742. Curran Associates, Inc., 2016.

\bibitem[Freund and Schapire(1997)]{freund97decision}
Yoav Freund and Robert~E Schapire.
\newblock A decision-theoretic generalization of on-line learning and an application to boosting.
\newblock \emph{Journal of Computer and System Sciences}, 55\penalty0 (1):\penalty0 119--139, 1997.

\bibitem[Freund and Schapire(1999)]{freund99adaptive}
Yoav Freund and Robert~E. Schapire.
\newblock Adaptive game playing using multiplicative weights.
\newblock \emph{Games and Economic Behavior}, 29\penalty0 (1):\penalty0 79--103, 1999.

\bibitem[Golowich et~al.(2020{\natexlab{a}})Golowich, Pattathil, and Daskalakis]{golowich20tight}
Noah Golowich, Sarath Pattathil, and Constantinos Daskalakis.
\newblock Tight last-iterate convergence rates for no-regret learning in multi-player games.
\newblock In \emph{Advances in Neural Information Processing Systems}, volume~33, pages 20766--20778. Curran Associates, Inc., 2020{\natexlab{a}}.

\bibitem[Golowich et~al.(2020{\natexlab{b}})Golowich, Pattathil, Daskalakis, and Ozdaglar]{golowich20last}
Noah Golowich, Sarath Pattathil, Constantinos Daskalakis, and Asuman Ozdaglar.
\newblock Last iterate is slower than averaged iterate in smooth convex-concave saddle point problems.
\newblock In \emph{Proceedings of Thirty Third Conference on Learning Theory}, volume 125, pages 1758--1784. PMLR, 2020{\natexlab{b}}.

\bibitem[Hart and Mas-Colell(2000)]{hart00simple}
Sergiu Hart and Andreu Mas-Colell.
\newblock A simple adaptive procedure leading to correlated equilibrium.
\newblock \emph{Econometrica}, 68\penalty0 (5):\penalty0 1127--1150, 2000.

\bibitem[Hart and Mas-Colell(2003)]{hart03uncoupled}
Sergiu Hart and Andreu Mas-Colell.
\newblock Uncoupled dynamics do not lead to {Nash} equilibrium.
\newblock \emph{American Economic Review}, 93\penalty0 (5):\penalty0 1830^^e2^^80^^931836, 2003.

\bibitem[Hsieh et~al.(2021)Hsieh, Antonakopoulos, and Mertikopoulos]{hsieh21adaptive}
Yu-Guan Hsieh, Kimon Antonakopoulos, and Panayotis Mertikopoulos.
\newblock Adaptive learning in continuous games: Optimal regret bounds and convergence to {Nash} equilibrium.
\newblock In \emph{Proceedings of Thirty Fourth Conference on Learning Theory}, volume 134, pages 2388--2422. PMLR, 2021.

\bibitem[Ito et~al.(2025)Ito, Luo, Tsuchiya, and Wu]{ito25instance}
Shinji Ito, Haipeng Luo, Taira Tsuchiya, and Yue Wu.
\newblock Instance-dependent regret bounds for learning two-player zero-sum games with bandit feedback.
\newblock In \emph{Proceedings of Thirty Eighth Conference on Learning Theory}, volume 291, pages 2858--2892. PMLR, 2025.

\bibitem[Koller et~al.(1996)Koller, Megiddo, and {von Stengel}]{koller96efficient}
Daphne Koller, Nimrod Megiddo, and Bernhard {von Stengel}.
\newblock Efficient computation of equilibria for extensive two-person games.
\newblock \emph{Games and Economic Behavior}, 14\penalty0 (2):\penalty0 247--259, 1996.

\bibitem[Li et~al.(2025)Li, Panpan, and Chen]{li25can}
Yuheng Li, Wang Panpan, and Haipeng Chen.
\newblock Can reinforcement learning solve asymmetric combinatorial-continuous zero-sum games?
\newblock In \emph{The Thirteenth International Conference on Learning Representations}, 2025.

\bibitem[Littlestone and Warmuth(1994)]{littlestone94weighted}
Nick Littlestone and Manfred~K Warmuth.
\newblock The weighted majority algorithm.
\newblock \emph{Information and computation}, 108\penalty0 (2):\penalty0 212--261, 1994.

\bibitem[Morav{\v{c}}{\'\i}k et~al.(2017)Morav{\v{c}}{\'\i}k, Schmid, Burch, Lis{\`y}, Morrill, Bard, Davis, Waugh, Johanson, and Bowling]{moravvcik17deepstack}
Matej Morav{\v{c}}{\'\i}k, Martin Schmid, Neil Burch, Viliam Lis{\`y}, Dustin Morrill, Nolan Bard, Trevor Davis, Kevin Waugh, Michael Johanson, and Michael Bowling.
\newblock Deepstack: Expert-level artificial intelligence in heads-up no-limit poker.
\newblock \emph{Science}, 356\penalty0 (6337):\penalty0 508--513, 2017.

\bibitem[Munos et~al.(2024)Munos, Valko, Calandriello, Gheshlaghi~Azar, Rowland, Guo, Tang, Geist, Mesnard, Fiegel, Michi, Selvi, Girgin, Momchev, Bachem, Mankowitz, Precup, and Piot]{munos24nash}
Remi Munos, Michal Valko, Daniele Calandriello, Mohammad Gheshlaghi~Azar, Mark Rowland, Zhaohan~Daniel Guo, Yunhao Tang, Matthieu Geist, Thomas Mesnard, C\^{o}me Fiegel, Andrea Michi, Marco Selvi, Sertan Girgin, Nikola Momchev, Olivier Bachem, Daniel~J Mankowitz, Doina Precup, and Bilal Piot.
\newblock {N}ash learning from human feedback.
\newblock In \emph{Proceedings of the 41st International Conference on Machine Learning}, volume 235, pages 36743--36768. PMLR, 2024.

\bibitem[Ouyang and Xu(2021)]{ouyang21lower}
Yuyuan Ouyang and Yangyang Xu.
\newblock Lower complexity bounds of first-order methods for convex-concave bilinear saddle-point problems.
\newblock \emph{Mathematical Programming}, 185\penalty0 (1):\penalty0 1--35, 2021.

\bibitem[Perolat et~al.(2022)Perolat, Vylder, Hennes, Tarassov, Strub, de~Boer, Muller, Connor, Burch, Anthony, McAleer, Elie, Cen, Wang, Gruslys, Malysheva, Khan, Ozair, Timbers, Pohlen, Eccles, Rowland, Lanctot, Lespiau, Piot, Omidshafiei, Lockhart, Sifre, Beauguerlange, Munos, Silver, Singh, Hassabis, and Tuyls]{perolat22mastering}
Julien Perolat, Bart~De Vylder, Daniel Hennes, Eugene Tarassov, Florian Strub, Vincent de~Boer, Paul Muller, Jerome~T. Connor, Neil Burch, Thomas Anthony, Stephen McAleer, Romuald Elie, Sarah~H. Cen, Zhe Wang, Audrunas Gruslys, Aleksandra Malysheva, Mina Khan, Sherjil Ozair, Finbarr Timbers, Toby Pohlen, Tom Eccles, Mark Rowland, Marc Lanctot, Jean-Baptiste Lespiau, Bilal Piot, Shayegan Omidshafiei, Edward Lockhart, Laurent Sifre, Nathalie Beauguerlange, Remi Munos, David Silver, Satinder Singh, Demis Hassabis, and Karl Tuyls.
\newblock Mastering the game of {Stratego} with model-free multiagent reinforcement learning.
\newblock \emph{Science}, 378\penalty0 (6623):\penalty0 990--996, 2022.

\bibitem[Rakhlin and Sridharan(2013)]{rakhlin13optimization}
Sasha Rakhlin and Karthik Sridharan.
\newblock Optimization, learning, and games with predictable sequences.
\newblock In \emph{Advances in Neural Information Processing Systems}, volume~26, pages 3066--3074. Curran Associates, Inc., 2013.

\bibitem[Swamy et~al.(2024)Swamy, Dann, Kidambi, Wu, and Agarwal]{swamy24minimaxlist}
Gokul Swamy, Christoph Dann, Rahul Kidambi, Steven Wu, and Alekh Agarwal.
\newblock A minimaximalist approach to reinforcement learning from human feedback.
\newblock In \emph{Proceedings of the 41st International Conference on Machine Learning}, volume 235, pages 47345--47377. PMLR, 2024.

\bibitem[Syrgkanis et~al.(2015)Syrgkanis, Agarwal, Luo, and Schapire]{syrgkanis15fast}
Vasilis Syrgkanis, Alekh Agarwal, Haipeng Luo, and Robert~E Schapire.
\newblock Fast convergence of regularized learning in games.
\newblock In \emph{Advances in Neural Information Processing Systems}, volume~28, pages 2989--2997. Curran Associates, Inc., 2015.

\bibitem[Tsuchiya et~al.(2025)Tsuchiya, Ito, and Luo]{tsuchiya25corrupted}
Taira Tsuchiya, Shinji Ito, and Haipeng Luo.
\newblock Corrupted learning dynamics in games.
\newblock In \emph{Proceedings of Thirty Eighth Conference on Learning Theory}, volume 291, pages 5506--5552. PMLR, 2025.

\bibitem[Washburn and Wood(1995)]{washburn95two}
Alan Washburn and Kevin Wood.
\newblock Two-person zero-sum games for network interdiction.
\newblock \emph{Operations Research}, 43\penalty0 (2):\penalty0 243--251, 1995.

\bibitem[Wei and Luo(2018)]{wei18more}
Chen-Yu Wei and Haipeng Luo.
\newblock More adaptive algorithms for adversarial bandits.
\newblock In \emph{Proceedings of the 31st Conference On Learning Theory}, volume~75, pages 1263--1291. PMLR, 2018.

\bibitem[Wei et~al.(2021)Wei, Lee, Zhang, and Luo]{wei21linear}
Chen-Yu Wei, Chung-Wei Lee, Mengxiao Zhang, and Haipeng Luo.
\newblock Linear last-iterate convergence in constrained saddle-point optimization.
\newblock In \emph{International Conference on Learning Representations}, 2021.

\bibitem[Wilder(2018)]{wilder18equilibrium}
Bryan Wilder.
\newblock Equilibrium computation and robust optimization in zero sum games with submodular structure.
\newblock \emph{Proceedings of the AAAI Conference on Artificial Intelligence}, 32\penalty0 (1), 2018.

\bibitem[Yoon and Ryu(2021)]{yoon21accelerated}
Taeho Yoon and Ernest~K Ryu.
\newblock Accelerated algorithms for smooth convex-concave minimax problems with {$O(1/k^2)$} rate on squared gradient norm.
\newblock In \emph{Proceedings of the 38th International Conference on Machine Learning}, volume 139, pages 12098--12109. PMLR, 2021.

\end{thebibliography}
\bibliographystyle{plainnat}

\newpage
\appendix

\section{Additional Related Work}\label{sec:related_work}
In this section, we discuss additional related work that could not be included in the main text.
In two-player zero-sum games, it was first pointed out by \citet{daskalakis11near} that a fast convergence rate of $\tilde{O}(1/T)$ is possible.
Subsequently, it was shown that both the optimistic Hedge algorithm and its generalized framework, optimistic follow-the-regularized-leader (OFTRL), can guarantee $\tilde{O}(1)$ social regret, which corresponds to an $\tilde{O}(1/T)$ convergence rate~\citep{rakhlin13optimization,syrgkanis15fast}.
Since then, achieving fast rates via such optimistic prediction has become a central approach when designing learning dynamics (\eg~\citealt{foster16learning,wei18more,chen20hedging,anagnostides22uncoupled}).
It is now known that such dynamics can guarantee an $O(1)$ individual regret upper bounds in two-player zero-sum games and an $O(\log T)$ bound in multiplayer general-sum games, ignoring dependencies other than on $T$.
It is worth noting that the optimistic Hedge algorithm does not always guarantee last-iterate convergence, but it has been shown to achieve last-iterate convergence under certain conditions~\citep{daskalakis19last,hsieh21adaptive,wei21linear}.

While a large body of work in learning in games has focused on deriving upper bounds, lower bounds remain relatively underexplored.
Several studies, including this work, investigated algorithm-dependent lower bounds.
\citet{syrgkanis15fast} considered a setting where the $x$-player employs the vanilla Hedge algorithm with an arbitrary learning rate, while the $y$-player plays a (pure) best response. 
For such a scenario, they showed that there exists an instance of learning in two-player zero-sum games in which the $x$-player must suffer $\sqrt{T}$ regret.
\citet{chen20hedging} showed that when both the $x$- and $y$-players use the vanilla Hedge with any learning rate, there exists a two-player \emph{general-sum} game in which at least one of the players incurs $\sqrt{T}$ regret.
In contrast, our work is the first to analyze regret lower bounds for \emph{optimistic} Hedge, to investigate their dependence on the numbers of actions $m$ and $n$, and to study the dynamic regret lower bounds.

Algorithm-dependent lower bounds on convergence rates have also been investigated in the context of last-iterate convergence.
For example, \citet{golowich20last} provided an $\Omega(1/\sqrt{T})$ lower bound on the last-iterate convergence rate for a class of algorithms that includes the extragradient method, and \citet{golowich20tight} established a similar lower bound for the optimistic gradient algorithm.
However, these results differ from ours not only in that they focus on the convergence rate of the last iterates, but also in that they assume an unconstrained setting.
In a related line of research, the fundamental limits of first-order methods have also been studied~\citep{ouyang21lower,yoon21accelerated}.
For a comprehensive overview of the literature on last-iterate convergence in the full-information (\ie~gradient feedback) setting, we refer the reader to \citet{cai25from} and the references therein.
It is worth noting that algorithm-independent analyses have also been explored in the literature.
The most relevant to our study is \citet{daskalakis11near}, who established an $\Omega(1/T)$ lower bound for strongly-uncoupled learning dynamics in two-player zero-sum games, which implies that learning dynamics based on optimistic Hedge are optimal up to a $\log(mn)$ factor.

\section{Omitted Details from \Cref{sec:regret_upper_bounds}}\label{app:proof_regret_upper_bounds}
This section provides deferred omitted details from \Cref{sec:regret_upper_bounds}.

\subsection{Regret Analysis of Optimistic Hedge (Proof of \Cref{lem:regxy_upper})}

Here we provide an analysis of optimistic Hedge.
In this subsection, we use the standard notation of online linear optimization.
Specifically, at each round $t = 1, \dots, T$, the player selects a point $w_t \in \calK$ from a convex feasible set $\calK \subseteq \R^d$, the environment chooses a loss vector $z_t \in \R^d$, and the player incurs a loss of $\inpr{w_t, z_t}$.
We use $D_\psi(v,w)$ to denote the Bregman divergence between $x$ and $y$ induced by a differentiable convex function $\psi$, that is,
$
  D_{\psi}(v, w) 
  = 
  \psi(v) - \psi(w) - \inpr{\nabla \psi(w), v -w}
$.

The following lemma provides a regret bound for the optimistic follow-the-regularized-leader (OFTRL), which generalizes the optimistic Hedge algorithm (adapted from \citealt[Lemma 16]{tsuchiya25corrupted}):
\begin{lemma}\label[lemma]{lem:oftrl_bound}
  Let $\calK \subseteq \R^d$ be a nonempty closed convex set.
  Suppose that a sequence of points $w_1, \dots, w_T \in \calK$ are selected by OFTRL,
  $w_t \in \argmin_{w \in \calK} \set{\inpr{w, m_t + \sum_{s=1}^{t-1} z_s} + \psi_t(w)}$, for each round $t \in [T]$.
  Then, for any $w^* \in \calK$, it holds that
  \begin{align}
    \sumT \inpr{w_t - w^*, z_t}
    &\leq
    \psi_{T+1}(w^*) - \psi_1(w_1)
    +
    \sumT \prn*{ \psi_t(w_{t+1}) - \psi_{t+1}(w_{t+1})}
    \nn 
    &\qquad+
    \sumT 
    \prn*{
      \inpr{w_t - w_{t+1}, z_t - m_t}
      -
      D_{\psi_t}(w_{t+1}, w_t)
    }
    +
    \inpr{w^* - w_{T+1}, m_{T+1}}
    \per 
    \label{eq:oftrl_bound}
  \end{align}
\end{lemma}

\Cref{lem:oftrl_bound} immediately yields the following regret upper bound: 

\begin{lemma}\label[lemma]{lem:oftrl_shannon}
  Let
  $\psi_t(w) = - \frac{1}{\eta_t} H(w)$ for
  $H(w) = \sum_{i=1}^d w(i) \log(1/w(i))$ be the negative Shannon entropy regularizer with nonincreasing learning rate $\eta_t$
  and 
  $w_t \in \argmin_{w \in \Delta_d} \set{\inpr{w, m_t + \sum_{s=1}^{t-1} z_s} + \psi_t(w)}$.
  Then, for any $w^* \in \Delta_d$ and $c_1, \dots, c_T > 0$, it holds that
  \begin{equation}
      \sumT \inpr{w_t - w^*, z_t}
    \leq
    \frac{\log d}{\eta_{T+1}}
    +
    \sumT
    \frac{\eta_t}{2 c_t} \nrm{z_t - m_t}_\infty^2 
    - 
    \sumT 
    \frac{1 - c_t}{2 \eta_t} \nrm{w_t - w_{t+1}}_1^2  
    + 2 \nrm{m_{T+1}}_\infty 
    \per 
    \n
    \end{equation}
\end{lemma}
This lemma generalizes \citet[Lemma 17]{tsuchiya25corrupted}.
Choosing $\eta_t = \eta$, $c_t = c$ for all $t \in [T]$ and $m_{T+1} = \zeros$ yields \Cref{lem:regxy_upper}.
\begin{proof}
We will upper bound the RHS of \Cref{eq:oftrl_bound} in \Cref{lem:oftrl_bound}.
From $\max_{w \in \Delta_d} H(w) \leq \log d$,
\begin{equation}
  \psi_{T+1}(w^*) 
  - \psi_1(w_1)
  +
  \sumT \prn*{ \psi_t(w_{t+1}) - \psi_{t+1}(w_{t+1})}
  \leq 
  \frac{\log d}{\eta_1}
  +
  \sumT \prn*{\frac{1}{\eta_{t+1}} - \frac{1}{\eta_t}} \log d
  =
  \frac{\log d}{\eta_{T+1}}
  \per
  \n
\end{equation}
Fix an arbitrary $c_t > 0$. Then, we also have
\begin{align}
  &
  \inpr{w_t - w_{t+1}, z_t - m_t}
  -
  D_{\psi_t}(x_{t+1}, x_t)
  \nn
  &=
  \inpr{w_t - w_{t+1}, z_t - m_t}
  -
  \frac{1}{\eta_t} D_{(- H)}(x_{t+1}, x_t)
  \nn
  &\leq 
  \nrm{w_t - w_{t+1}}_1 \nrm{z_t - m_t}_\infty
  -
  \frac{1}{2 \eta_t} \nrm{w_t - w_{t+1}}_1^2
  \nn 
  &=
  \nrm{w_t - w_{t+1}}_1 \nrm{z_t - m_t}_\infty
  -
  \frac{c_t}{2\eta_t} \nrm{w_t - w_{t+1}}_1^2
  -
  \frac{1 - c_t}{2\eta_t} \nrm{w_t - w_{t+1}}_1^2
  \nn 
  &\leq 
  \frac{\eta_t}{2 c_t} \nrm{z_t - m_t}_\infty^2 
  - \frac{1 - c_t}{2 \eta_t} \nrm{w_t - w_{t+1}}_1^2
  \com 
  \n
\end{align}
where the first inequality follows from H\"older's inequality and the fact that the function $(- H)$ is $1$-strongly convex with respect to $\nrm{\cdot}_1$, and the last inequality follows by considering the worst-case with respect to $\nrm{w_t - w_{t+1}}_1$ in the first two terms.
Combining \Cref{lem:oftrl_bound} with the above two inequalities completes the proof.
\end{proof}

\subsection{Proof of \Cref{thm:upper_bounds}}
Here we provide the proof of \Cref{thm:upper_bounds}.
\begin{proof}[Proof of \Cref{thm:upper_bounds}]
Summing the regret upper bounds in \Cref{lem:regxy_upper} and using 
$\nrm{g_1 - g_0}_\infty \leq 1$, $\nrm{\ell_1 - \ell_0}_\infty \leq 1$,
$\nrm{g_t - g_{t-1}}_\infty = \nrm{A (y_t - y_{t-1})}_\infty \leq \nrm{y_t - y_{t-1}}_1$ and $\nrm{\ell_t - \ell_{t-1}}_\infty \leq \nrm{x_t - x_{t-1}}_1$ that hold for all $t \in [T]$,
we have
\begin{align}
  &
  \Reg_T^x + \Reg_T^y
  \nn
  &\leq
  \frac{\log m}{\eta}
  +
  \frac{\eta}{2 c}
  +
  \frac{\log n}{\eta'}
  +
  \frac{\eta'}{2 c'}
  +
  \prn*{\frac{\eta'}{2 c'} - \frac{1 - c}{2 \eta}} 
  \sum_{t=2}^T \nrm{x_t - x_{t-1}}_1^2 
  + 
  \prn*{\frac{\eta}{2 c} - \frac{1 - c'}{2 \eta'}} 
  \sum_{t=2}^T \nrm{y_t - y_{t-1}}_1^2
  \nn
  &=
  \Omega(\eta, \eta', c, c')
  +
  \prn*{\frac{\eta'}{2 c'} - \frac{1 - c}{2 \eta}} 
  \sum_{t=2}^T \nrm{x_t - x_{t-1}}_1^2 
  + 
  \prn*{\frac{\eta}{2 c} - \frac{1 - c'}{2 \eta'}} 
  \sum_{t=2}^T \nrm{y_t - y_{t-1}}_1^2
  \com
  \label{eq:reg_sum_comp}
\end{align}
where we recall 
\begin{equation}
  \Omega(\eta, \eta', c, c')
  =
  \omega(\eta, c)
  +
  \omega'(\eta', c')
  \com\quad
  \omega(\eta, c)
  =
  \frac{\log m}{\eta}
  +
  \frac{\eta}{2 c}
  \com\quad
  \omega'(\eta', c')
  =
  \frac{\log n}{\eta'}
  +
  \frac{\eta'}{2 c'}
  \per
  \n
\end{equation}

Note that when $\eta \eta' < c' (1 - c)$, we have
$\frac{\eta'}{2 c'} - \frac{1 - c}{2 \eta} < 0$
and 
when $\eta \eta' < c (1 - c')$ we have 
$\frac{\eta}{2 c} - \frac{1 - c'}{2 \eta'} < 0$.
Hence, when $\eta \eta' \leq c' (1 - c)$ and $\eta \eta' < c (1 - c')$, from \Cref{eq:reg_sum_comp} and the fact that $\SReg_T \geq 0$, we have
\begin{equation}
  \sum_{t=2}^T \nrm{y_t - y_{t-1}}_1^2 
  \leq
  \frac{1}{\frac{1 - c'}{2 \eta'} - \frac{\eta}{2 c}}
  \cdot
  \omega(\eta, \eta', c, c')
  \per
  \n
\end{equation}
Hence, combining \Cref{lem:regxy_upper} with 
$\nrm{g_t - g_{t-1}}_\infty \leq \nrm{y_t - y_{t-1}}_1$
and the last inequality, we obtain 
\begin{equation}
  \Reg_T^x
  \leq
  \frac{\log m}{\eta}
  +
  \frac{\eta}{2 c}
  +
  \frac{\eta}{2 c} \sum_{t=2}^T \nrm{y_t - y_{t-1}}_\infty^2
  \leq
  \omega(\eta, c)
  +
  \frac{\frac{\eta}{2 c}}{ \frac{1 - c'}{2 \eta'} - \frac{\eta}{2 c} }
  \cdot
  \Omega(\eta, \eta', c, c')
  =
  f(\eta, \eta', c, c')
  \per
  \n
\end{equation}

Similarly, if $\eta \eta' < c' (1 - c)$ and $\eta \eta' \leq c (1 - c')$, from \Cref{eq:reg_sum_comp} and the fact that $\SReg_T \geq 0$ we have
\begin{equation}
  \sum_{t=2}^T \nrm{x_t - x_{t-1}}_1^2 
  \leq
  \frac{1}{\frac{1 - c}{2 \eta} - \frac{\eta'}{2 c'}}
  \cdot
  \omega(\eta, \eta', c, c')
  \per
  \n
\end{equation}
Hence, combining \Cref{lem:regxy_upper} with
$\nrm{\ell_t - \ell_{t-1}}_\infty \leq \nrm{x_t - x_{t-1}}_1$
and the last inequality, we obtain
\begin{equation}
  \Reg_T^y
  \leq
  \frac{\log n}{\eta'}
  +
  \frac{\eta'}{2 c'}
  +
  \frac{\eta'}{2 c'} \sum_{t=2}^T \nrm{x_t - x_{t-1}}_\infty^2
  \leq
  \omega'(\eta', c')
  +
  \frac{\frac{\eta'}{2 c'}}{\frac{1 - c}{2 \eta} - \frac{\eta'}{2 c'}}
  \cdot
  \Omega(\eta, \eta', c, c')
  =
  g(\eta, \eta', c, c')
  \com
  \n
\end{equation}
This completes the proof.
\end{proof}

\subsection{Social Regret Analysis Deferred from \Cref{subsec:social_upper}}
Here we first provide the proof of \Cref{lem:optimal_large_omega}, which will be used to obtain the tight upper bound on the social regret.
We then provide the proof of \Cref{thm:optimal_sreg_su}.

\subsubsection{Proof of \Cref{lem:optimal_large_omega} (cardinality-aware case)}
Recall that \Cref{lem:optimal_large_omega} is for cardinality-aware strongly-uncoupled learning dynamics,
and thus we can choose $\lambda$ and $\lambda'$ by considering the RHS of the following inequality:
\begin{equation}
  \SReg_T
  \leq
  \inf_{\lambda \in \Lambda}
  \Omega(\eta, \eta', c, c')
  \com\quad
  \Omega(\eta, \eta', c, c')
  =
  \frac{M}{\eta}
  +
  \frac{\eta}{2 c}
  +
  \frac{N}{\eta'}
  +
  \frac{\eta'}{2 c'}
  \com
  \label{eq:sreg_opt_prob_casu}
\end{equation}
where we recall that
$
  \Lambda
  =
  \set{
    \lambda = (\eta, \eta', c, c') \in \Rp^4
    \colon
    \eta \eta' \leq c' (1 - c),\, \eta \eta' \leq c (1 - c') 
  }
$, $M = \log m$, and $N = \log n$.

\begin{proof}[Proof of \Cref{lem:optimal_large_omega}]
From the constraints that $\eta, \eta' > 0$ and $\eta \eta' \leq c' (1 - c),\, \eta \eta' \leq c (1 - c')$, we have $c, c' \in (0, 1)$.
Hence, the constraints
$\eta \eta' \leq c' (1 - c)$ and $\eta \eta' \leq c (1 - c')$
can be rewritten as
$\frac{\eta}{c} \cdot \frac{\eta'}{c'} \leq \frac{1}{c} - 1$
and
$\frac{\eta}{c} \cdot \frac{\eta'}{c'} \leq \frac{1}{c'} - 1$, respectively.

Now we consider the following change of variables:
\begin{equation}
  a = \frac{\eta}{c}
  \com\quad
  a' = \frac{\eta'}{c'}
  \com\quad
  b = \frac{1}{c} - 1
  \com\quad
  b' = \frac{1}{c'} - 1
  \per
  \label{eq:def_ab}
\end{equation}
Note that this is a bijective transformation, 
and we have
\begin{equation}
  c = \frac{1}{1 + b} \in (0 ,1)
  \com\quad
  c' = \frac{1}{1 + b'} \in (0 ,1)
  \com\quad
  \eta = ac
  \com\quad
  \eta' = a' c' 
  \per
  \label{eq:c_to_ab}
\end{equation}
Then the constraints $\eta \eta' \leq c' (1 - c)$ and $\eta \eta' \leq c (1 - c')$ can be rewritten as
$
  a a' \leq b
$
and
$
a a' \leq b'
$,
respectively,
and the RHS of the inequality in \Cref{eq:sreg_opt_prob_casu} can rewritten as
\begin{equation}
  \inf_{a,a',b,b' > 0 \colon a a' \leq b, a a' \leq b'} 
  \Omega(a,a',b,b')
  \com\quad
  \Omega(a,a',b,b')
  =
  (1 + b) \frac{M}{a}
  +
  \frac{a}{2}
  +
  (1 + b')
  \frac{N}{a'}
  +
  \frac{a'}{2}
  \com
  \n
\end{equation}
where we abuse the notation of $\Omega$.
The rewritten function $\Omega$ is monotonically increasing with respect to $b$ and $b'$.
Hence the optimal choices of $b$ and $b'$ is $b = b' = a a'$ and in this case, we have
\begin{equation}
  c = c' = \frac{1}{1 + a a'}
  \com\quad
  \eta \eta' 
  = 
  \frac{a a'}{\prn{1 + a a'}^2}
  =
  c' (1 - c)
  =
  c (1 - c')
  \com
  \n
\end{equation}
and
\begin{equation}
  \Omega(a,a',b,b')
  =
  (1 + aa') \prn*{ \frac{M}{a} + \frac{N}{a'} } 
  +
  \frac{a}{2}
  +
  \frac{a'}{2}
  =
  \prn*{
    \frac{M}{a}
    +
    \prn*{N + \frac12} a
  }
  +
  \prn*{
    \frac{N}{a'}
    +
    \prn*{M + \frac12} a'
  }
  \per
  \n
\end{equation}
From the AM--GM inequality, choosing
\begin{equation}
  a
  =
  \sqrt{\frac{M}{N + \frac12}}
  \com\quad
  a'
  =
  \sqrt{\frac{N}{M + \frac12}}
  \com
  \n
\end{equation}
gives the minimum value of $\Omega$ and
its optimal value is $2 \sqrt{M \prn*{N + \frac12}} + 2 \sqrt{N \prn*{M + \frac12}}$.
From \Cref{eq:c_to_ab}, the optimal parameters of \Cref{eq:sreg_opt_prob_casu} are given by
\begin{equation}
  c \!=\! c'
  \!=\!
  \frac{\sqrt{\! \prn*{M \!+\! \frac12} \prn*{N \!+\! \frac12}}}{\sqrt{\! \prn*{M \!+\! \frac12} \! \prn*{N \!+\! \frac12}} \!+\! \sqrt{M N}}
  \com\
  \eta
  \!=\!
  \frac{\sqrt{M \prn*{M \!+\! \frac12}}}{\sqrt{\!\prn*{M \!+\! \frac12} \! \prn*{N \!+\! \frac12}} \!+\! \sqrt{M N}}
  \com\
  \eta'
  \!=\!
  \frac{\sqrt{N \prn*{N \!+\! \frac12}}}{\sqrt{\! \prn*{M \!+\! \frac12} \! \prn*{N \!+\! \frac12}} \!+\! \sqrt{M N}}
  \com
  \n
\end{equation}
which are indeed elements in the feasible set $\Lambda$, and we have completed the proof of \Cref{lem:optimal_large_omega}.
\end{proof}

\subsubsection{Proof of \Cref{thm:optimal_sreg_su} (cardinality-unaware case)}
We next provide the proof of \Cref{thm:optimal_sreg_su}.
Under strongly-uncoupled learning dynamics without cardinality-awareness, the learning rates $\eta, \eta' > 0$ cannot be chosen as functions of $M = \log m$ and $N = \log n$.
Note, however, that the parameters $c, c' > 0$ are not algorithm-dependent variables and thus may depend on $M$ and $N$.

\begin{proof}[Proof of \Cref{thm:optimal_sreg_su}]
Define
\begin{equation}
  \Lambda(\eta, \eta')
  =
  \set{
    (c, c') \in \Rp^2
    \colon
    \eta \eta' \leq c' (1 - c),\, \eta \eta' \leq c (1 - c') 
  }
  \per
  \n
\end{equation}
Then, we will show that
\begin{equation}
  \min_{(c, c') \in \Lambda(\eta, \eta')}
  \set*{
    \frac{\eta}{2 c}
    +
    \frac{\eta'}{2 c'}
  }
  =
  \frac{\eta + \eta'}{1 + \sqrt{1 - 4 \eta \eta'}}
  \com
  \n
\end{equation}
and the optimal $c, c' \in \Lambda(\eta, \eta')$ achieving the minimum are given by 
$c = c' = \frac{1 + \sqrt{1 - 4 \eta \eta'}}{2}$.
To prove this, from the KKT condition,
letting
$
\calL(c, c', \mu_1, \mu_2)
=
\frac{\eta}{2 c}
+
\frac{\eta'}{2 c'}
+
\mu_1 \prn{\eta \eta' - c' (1 - c)}
+
\mu_2 \prn{\eta \eta' - c (1 - c')}
$, we have
\begin{equation}
  \begin{split}  
    &
    \frac{\partial \calL}{\partial c}
    =
    - \frac{\eta}{2 c^2}
    +
    \mu_1 c'
    -
    \mu_2 (1 - c')
    = 
    0
    \com\quad
    \frac{\partial \calL}{\partial c'}
    =
    - \frac{\eta'}{2 c'^2}
    -
    \mu_1 (1 - c)
    +
    \mu_2 c
    =
    0
    \com\quad
    \\
    &
    \mu_1 \prn{\eta \eta' - c' (1 - c)} = 0
    \com
    \quad
    \mu_2 \prn{\eta \eta' - c (1 - c')} = 0
    \com\quad
    \mu_1, \mu_2 \geq 0
    \per
  \end{split}
  \label{eq:kkt_social_su}
\end{equation}
From the third and fourth equalities in \Cref{eq:kkt_social_su}, we claim that $\mu_1, \mu_2 > 0$.
Indeed, if $\mu_1 = 0$, then the first equality in \Cref{eq:kkt_social_su} gives
$
- \frac{\eta}{2 c^2}
-
\mu_2 (1 - c')
= 
0
$, which is impossible since $c \in (0, 1)$ and $\mu_2 \geq 0$.
Similarly, if $\mu_2 = 0$, then the second equality in \Cref{eq:kkt_social_su} gives
$
- \frac{\eta'}{2 c'^2}
-
\mu_1 (1 - c)
=
0
$, which is a contradiction.
Hence, from $\mu_1, \mu_2 > 0$, we have $\eta \eta' = c' (1 - c) = c(1 - c')$.
From these two equalities, we have $c = c'$.
Hence, $c (1 - c) = \eta \eta'$ and thus $c = c' = \frac{1 + \sqrt{1 - 4 \eta \eta'}}{2}$.

From the above observation, it suffices to choose absolute constants $\eta,\eta' > 0$ (independent of $M,N$) to minimize
\begin{equation}
  \min_{(c, c') \in \Lambda(\eta, \eta')}
  \Omega(\eta, \eta', c, c')
  =
  \frac{M}{\eta}
  +
  \frac{N}{\eta'}
  +
  \frac{\eta + \eta'}{1 + \sqrt{1 - 4 \eta \eta'}}
  \eqqcolon
  F(\eta, \eta')
  \per
  \n
\end{equation}
A natural approach is to minimize a linear upper bound that holds for all $M,N>0$:
\begin{equation}
  F(\eta, \eta')
  \leq
  S(\eta, \eta') (M + N)
  +
  \frac{\eta + \eta'}{1 + \sqrt{1 - 4 \eta \eta'}}
  \com\quad
  S(\eta, \eta') = \max\set*{\frac{1}{\eta},\, \frac{1}{\eta'}}
  \per
  \n
\end{equation}
The slope $S(\eta, \eta')$ is minimized when $\eta = \eta'$.
Since $\eta \eta' \leq \max_{c, c' \in [0, 1]} \min\set{c' (1 - c), c (1 - c')} = 1/4$,
choosing $\eta = \eta' = 1/2$ and thus $c = c' = 1/2$ are optimal choices.
Therefore, with these choices of $\eta, \eta'$, for all $M, N > 0$ we have
\begin{equation}
  F(\eta, \eta')
  \leq
  2 (M + N) + 1
  \com
  \n
\end{equation}
which leads to the desired social regret upper bound under strongly-uncoupled learning dynamics.
\end{proof}

\subsection{Individual Regret Analysis Deferred from \Cref{subsec:indiv_upper}}
Here we provide deferred details from \Cref{subsec:indiv_upper}.
Recall that, as defined in \Cref{thm:upper_bounds}, $f$ and $g$ are given by
\begin{equation}
  \begin{split}    
    f(\eta, \eta', c, c')
    &\!=\!
    \omega(\eta, c)
    \!+\!
    \frac{\frac{\eta}{2 c}}{ \frac{1 - c'}{2 \eta'} \!-\! \frac{\eta}{2 c} }
    \Omega(\eta, \eta', c, c')
    =
    \frac{\log m}{\eta}
    \!+\!
    \frac{\eta}{2 c}
    \!+\!
    \frac{\frac{\eta}{2 c}}{ \frac{1 - c'}{2 \eta'} - \frac{\eta}{2 c} }
    \prn*{
      \frac{\log m}{\eta}
      \!+\!
      \frac{\log n}{\eta'}
      \!+\!
      \frac{\eta}{2 c}
      \!+\!
      \frac{\eta'}{2 c'}
    }
    \com
    \\
    g(\eta, \eta', c, c')
    &\!=\!
    \omega'(\eta', c')
    \!+\!
    \frac{\frac{\eta'}{2 c'}}{\frac{1 - c}{2 \eta} \!-\! \frac{\eta'}{2 c'}}
    \Omega(\eta, \eta', c, c')
    \!=\!
    \frac{\log n}{\eta'}
    \!+\!
    \frac{\eta'}{2 c'}
    \!+\!
    \frac{\frac{\eta'}{2 c'}}{\frac{1 - c}{2 \eta} - \frac{\eta'}{2 c'}}
    \prn*{
      \frac{\log m}{\eta}
      \!+\!
      \frac{\log n}{\eta'}
      \!+\!
      \frac{\eta}{2 c}
      \!+\!
      \frac{\eta'}{2 c'}
    }
    \com
  \end{split}
  \label{eq:redef_fg_arg_eta_c}
\end{equation}
where $\omega$, $\omega'$, and $\Omega$ are defined in \Cref{eq:def_omega}.

\subsubsection{Common Analysis}\label{app:common_args_as}
We introduce the variables $s, s' \geq 0$ so that the optimization problem for $f, g$ in \Cref{eq:redef_fg_arg_eta_c} can be equivalently expressed as an optimization problem over $(a, a', s, s')$:
\begin{equation}
  s = \frac{b}{a} - a' = \frac{b - a a'}{a} \geq 0
  \com\quad
  s' = \frac{b'}{a'} - a = \frac{b' - a a'}{a'} \geq 0
  \com
  \label{eq:def_s}
\end{equation}
where $(a, a', b, b')$ are defined in \Cref{eq:def_ab}.
Then, we have
\begin{align}
  c
  =
  \frac{1}{1 + b}
  =
  \frac{1}{1 + a (a' + s)}
  \com\quad
  c
  =
  \frac{1}{1 + a'(a + s')}
  \com\quad
  \eta
  =
  c a
  =
  \frac{a}{1 + a (a' + s)}
  \com\quad
  \eta'
  =
  \frac{a'}{1 + a'(a + s')}
  \per
  \label{eq:a_s_to_c_eta}
\end{align}
Using $(a, a', s, s')$, we can rewrite $\omega$, $\omega'$, and $\Omega$ in \Cref{eq:def_omega} as (we abuse notation of $\omega$, $\omega'$, and $\Omega$ again)
\begin{align}
  &
  \omega(a, a', s)
  =
  \frac{M}{\eta}
  +
  \frac{\eta}{2 c}
  =
  \frac{M}{a}
  +
  (a' + s) M
  +
  \frac{a}{2}
  \com\quad
  \omega'(a, a', s')
  =
  \frac{N}{\eta'}
  +
  \frac{\eta'}{2 c'}
  =
  \frac{N}{a'}
  +
  (a + s') N
  +
  \frac{a'}{2}
  \com
  \nn
  &
  \Omega(a, a', s, s')
  =
  h(a, a')
  +
  s M + s' N
  \com
  \n
\end{align}
where we defined
\begin{equation}
  h(a, a')
  \coloneqq
  \frac{M}{a}
  +
  a' M
  +
  \frac{N}{a'}
  +
  a N
  +
  \frac{a}{2}
  +
  \frac{a'}{2}
  =
  \frac{M}{a}
  +
  \prn*{N + \frac12} a
  +
  \frac{N}{a'}
  +
  \prn*{M + \frac12} a'
  \per
  \label{eq:def_h}
\end{equation}
We also have
\begin{equation}
  \frac{1 - c'}{2 \eta'} - \frac{\eta}{2 c}
  =
  \frac12 \prn*{ \frac{(1 - c') / c'}{\eta' / c'} - a}
  =
  \frac12 \prn*{ \frac{b'}{a'} - a}
  =
  \frac{s'}{2}
  \com\quad
  \frac{1 - c}{2 \eta} - \frac{\eta'}{2 c'}
  =
  \frac{s}{2}
  \com
  \n
\end{equation}
and thus $f$ and $g$ in \Cref{eq:redef_fg_arg_eta_c} can be rewritten as
\begin{equation}
  \begin{split}
    f(a, a', s, s')
    &=
    \prn*{
      \frac{M}{a}
      +
      a' M
      +
      \frac{a}{2}
      +
      a N
    }
    +
    s M
    +
    \frac{a}{s'}
    \prn*{
      h(a, a')
      +
      s M 
    }
    \com
    \\
    g(a, a', s, s')
    &=
    \prn*{
      \frac{N}{a'}
      +
      a N
      +
      \frac{a'}{2}
      +
      a' M
    }
    +
    s' N
    +
    \frac{a'}{s}
    \prn*{
      h(a, a')
      + 
      s' N
    }
    \per
  \end{split}
  \label{eq:def_fg_arg_as}
\end{equation}
where we replace the arguments $(\eta, \eta', c, c')$ of $f, g$ with $(a, a', s, s')$ by abuse of notation.

\subsubsection{Extreme Cases}
We supplement the discussion of the extreme cases, where the goal is to minimize only one player's individual regret, which was omitted in the main text.
We considered the setting where the $x$-player uses optimistic Hedge (with a learning rate of either $\eta = \sqrt{M / (N + 1/2)}$ in the cardinality-aware case or $\eta = 1$ in the cardinality-unaware case), while the $y$-player plays the uniform strategy at every round.
There, we showed that by the asymptotic argument (that is in fact not applicable in this limiting scenario) it holds that $\Reg_T^x \leq \sqrt{M \prn*{N + 1/2}}$ in \Cref{eq:x_best_indiv_casu} for the cardinality-aware case and $\Reg_T^x \leq M + N + 1/2$ in \Cref{eq:x_best_indiv_su} for the cardinality-unaware case.
These bounds (or their improved bounds) can be derived directly from a direct analysis.
Specifically, combining the $x$-player's regret upper bound in \Cref{lem:regxy_upper} with the fact that $g_t = g_{t-1} = A (\frac{1}{n} \ones)$ for all $t \in [T]$, we have
\begin{align}
  \Reg_T^x
  &\leq
  \inf_{c > 0} \set*{
    \frac{\log m}{\eta}
    +
    \frac{\eta}{2 c} \sumT \nrm{g_t - g_{t-1}}_\infty^2
    -
    \frac{1 - c}{2 \eta} \sum_{t=2}^T \nrm{x_t - x_{t-1}}_1^2
  }
  \nn
  &\leq
  \frac{\log m}{\eta}
  +
  \frac{\eta}{2}
  =
  \begin{cases}
    \frac{3}{2} \sqrt{M \prn*{N + \frac12}} & \mbox{cardinality-aware case} \com \\
    M + \frac12 & \mbox{cardinality-unaware case} \com
  \end{cases}
  \n
\end{align}
where we chose $c = 1$ and used $\nrm{g_1 - g_0}_\infty \leq 1$.

\subsubsection{Upper Bounding the Maximum of the Individual Regrets}

Here we provide the omitted details to upper bound the maximum of the individual regrets provided in \Cref{lem:optimal_J_gamma} and \Cref{thm:optimal_max_indivreg_casu,thm:optimal_max_indivreg_su}.
From the analysis in \Cref{app:common_args_as}, we can rewrite $J_\gamma(\eta, \eta', c, c')$ in \Cref{eq:def_Jgamma} as
\begin{align}
  \begin{split}    
    J_\gamma(a, a', s, s')
    &=
    \gamma
    \prn*{
      \frac{M}{a}
      +
      a' M
      +
      \frac{a}{2}
      +
      a N
    }
    +
    (1 - \gamma)
    \prn*{
      \frac{N}{a'}
      +
      a N
      +
      \frac{a'}{2}
      +
      a' M
    }
    \\
    &\qquad
    +
    \gamma
    s M
    +
    \gamma
    \frac{a}{s'}
    \prn*{
      h(a, a')
      +
      s M 
    }
    +
    (1 - \gamma)
    s' N
    +
    (1 - \gamma)
    \frac{a'}{s}
    \prn*{
      h(a, a')
      + 
      s' N
    }
    \com
  \end{split}
  \label{eq:Jgamma_arg_as}
\end{align}
where we replaced the arguments $(\eta, \eta', c, c')$ with $(a, a', s, s')$ by abuse of notation.

\paragraph{Cardinality-aware case}
As discussed in the main text, in the cardinality-aware case it is difficult to compute a closed-form expression for the exact minimum of the individual regret or the minimizer that attains it.
To address this, we will derive an upper bound of $\max\set{f, g}$.
Specifically, we will focus on the case of $\gamma = 1/2$, which is for \Cref{lem:optimal_J_gamma} and \Cref{thm:optimal_max_indivreg_casu}.

\begin{proof}[Proof of \Cref{lem:optimal_J_gamma}]
  
We consider the case where $s, s' \geq 0$ are expressed as
$
  s = \theta a'
$
and
$
  s' = \theta a
$
for some $\theta > 0$.
In this case, we can rewrite $J_\gamma$ in \Cref{eq:Jgamma_arg_as} as
\begin{align}
  J_\gamma(a, a', s, s')
  &=
  \frac12
  \prn*{
    \frac{M}{a}
    +
    a' M
    +
    \frac{a}{2}
    +
    a N
  }
  +
  \frac12
  \prn*{
    \frac{N}{a'}
    +
    a N
    +
    \frac{a'}{2}
    +
    a' M
  }
  \nn
  &\qquad
  +
  \frac12 \theta a' M
  +
  \frac{1}{2 \theta}
  \prn*{
    h(a, a')
    +
    \theta a' M 
  }
  +
  \frac12
  \theta a N
  +
  \frac{1}{2 \theta}
  \prn*{
    h(a, a')
    + 
    \theta a N
  }
  \nn
  &=
  \frac{1}{2 a}
  \prn*{
    M + \frac{2 M}{\theta}
  }
  +
  \frac{a}{2}
  \prn*{
    \frac12 + 2 N + \frac{2 (N + 4)}{\theta} + \theta N + N
  }
  \nn
  &\qquad+
  \frac{1}{2 a'}
  \prn*{
    N + \frac{2 N}{\theta}
  }
  +
  \frac{a'}{2}
  \prn*{
    \frac12 + 2 M + \frac{2 (M + 4)}{\theta} + \theta M + M
  }
  \com
\end{align}
where $h$ is given in \Cref{eq:def_h}.
Then, we can evaluate $  \inf_{\lambda \in \Lambda} J_{1/2}(\lambda) = \inf_{a, a', s, s' > 0} J_{1/2}(a, a', s, s')$ as
\begin{align}
  \inf_{\lambda \in \Lambda}
  J_{1/2}(\lambda)
  &\leq
  \inf_{\theta > 0}
  \set*{
  \sqrt{
    M
    \prn*{
      1 + \frac{2}{\theta}
    }
    \prn*{
      \prn*{\theta + 3 + \frac{2}{\theta}} N + \frac{8}{\theta} + \frac12
    }
  }
  +
  \sqrt{
    N
    \prn*{
      1 + \frac{2}{\theta}
    }
    \prn*{
      \prn*{\theta + 3 + \frac{2}{\theta}} M + \frac{8}{\theta} + \frac12
    }
  }
  }
  \nn
  &\leq
  \sqrt{\frac{5}{3} M \cdot \prn*{\frac{20}{3} N + \frac{19}{6}}}
  +
  \sqrt{\frac{5}{3} N \cdot \prn*{\frac{20}{3} M + \frac{19}{6}}}
  \nn
  &\leq
  \sqrt{\frac{5}{3} M \cdot \frac{20}{3} \prn*{ N + \frac12}}
  +
  \sqrt{\frac{5}{3} N \cdot \frac{20}{3} \prn*{ M + \frac12}}
  \nn
  &=
  \frac{10}{3} \prn*{\sqrt{M \prn*{N + \frac12}} + \sqrt{N \prn*{M + \frac12}}}
  \com
  \n
\end{align}
where in the first inequality, we chose
\begin{equation}
  a
  =
  \sqrt{
    \frac{
      M + \frac{2 M}{\theta}
    }{
      \frac23 + (\theta + 3) N + \frac{2 (N + 4)}{\theta}
    }
  }
  \com\quad
  a'
  =
  \sqrt{
    \frac{
      N + \frac{2 N}{\theta}
    }{
      \frac23 + (\theta + 3) M + \frac{2 (M + 4)}{\theta}
    }
  }
  \com
  \n
\end{equation}
(here we chose the first term in the denominator inside the square roots of $a$ and $a'$ to be $2/3$ instead of $1/2$, which is suboptimal but simplifies the resulting expressions)
and in the second inequality we chose $\theta = 3$.
Therefore,
\begin{equation}
  \inf_{\lambda \in \Lambda} \max\set{f(\lambda), g(\lambda)}
  \leq
  2
  \cdot 
  \inf_{\lambda \in \Lambda} \frac{f(\lambda) + g(\lambda)}{2}
  =
  2
  \cdot 
  \inf_{\lambda \in \Lambda} J_{1/2}(\lambda)
  \leq
  \frac{20}{3} \prn*{\sqrt{M \prn*{N + \frac12}} + \sqrt{N \prn*{M + \frac12}}}
  \per
  \n
\end{equation}
The corresponding $a$ and $a'$ achieving the above bound is
\begin{equation}
  a
  =
  \sqrt{
    \frac{\frac{5}{3} M}{\frac23 + 6 N + \frac{2}{3} (N + 4)}
  }
  =
  \sqrt{
    \frac{\frac{5}{3} M}{\frac{20}{3} N + \frac{10}{3}}
  }
  =
  \frac12
  \sqrt{
    \frac{M}{N + \frac12}
  }
  \com\quad
  a'
  =
  \frac12
  \sqrt{\frac{N}{M + \frac12}}
  \com
  \n
\end{equation}
and thus corresponding $\eta, \eta' > 0$ are given by
\begin{align}
  \eta
  &=
  \frac{a}{1 + a(a' + s)}
  =
  \frac{a}{1 + a(a' + \theta a')}
  =  
  \frac{
    \frac{1}{2} \sqrt{\frac{M}{N + 1/2}}
  }{
    1 + \sqrt{\frac{M N}{(M + 1/2)(N + 1/2)}}
  }
  =
  \frac12
  \frac{\sqrt{M (M + 1/2)}}{\sqrt{(M + 1/2)(N + 1/2)} + \sqrt{M N}}
  \com
  \nn
  \eta'
  &=
  \frac12\frac{\sqrt{N (N + 1/2)}}{\sqrt{(M + 1/2)(N + 1/2)} + \sqrt{M N}}
  \per
  \n
\end{align}
This completes the proof.
\end{proof}

\paragraph{Cardinality-unaware case}
We next provide the proof of \Cref{thm:optimal_max_indivreg_su}, which is for the cardinality-unaware case.

\begin{proof}[Proof of \Cref{thm:optimal_max_indivreg_su}]
Recall the rewritten forms of $f$ and $g$ in \Cref{eq:def_fg_arg_as}:
\begin{equation}
  \begin{split}
    f(a, a', s, s')
    &=
    \prn*{
      \frac{M}{a}
      +
      a' M
      +
      \frac{a}{2}
      +
      a N
    }
    +
    s M
    +
    \frac{a}{s'}
    \prn*{
      h(a, a')
      +
      s M 
    }
    \com
    \nn
    g(a, a', s, s')
    &=
    \prn*{
      \frac{N}{a'}
      +
      a N
      +
      \frac{a'}{2}
      +
      a' M
    }
    +
    s' N
    +
    \frac{a'}{s}
    \prn*{
      h(a, a')
      + 
      s' N
    }
    \com
  \end{split}
\end{equation}
where we recall
$
  h(a, a')
  =
  \frac{M}{a}
  +
  a' M
  +
  \frac{N}{a'}
  +
  a N
  +
  \frac{a}{2}
  +
  \frac{a'}{2}
$
as defined in \Cref{eq:def_h}.

As in the analysis of  \Cref{thm:optimal_sreg_su}, we minimize the worst-case coefficients of $f$ and $g$ with respect to $M$ and $N$.
In particular, we define the coefficients of $f$ and $g$ with respect to $M$ and $N$ as follows:
\begin{equation}
  \begin{split}
    \kappa_M^f
    &=
    \frac{1}{a} + a' + s + \frac{a}{s'} \prn*{\frac{1}{a} + a' + s}
    =
    \prn*{1 + \frac{a}{s'}} \prn*{\frac{1}{a} + a' + s}
    \com\quad
    \kappa_N^f
    =
    a + \frac{a}{s'} \prn*{\frac{1}{a'} + a}
    =
    \frac{a}{s'} \prn*{\frac{1}{a'} + a + s'}
    \com
    \\
    \kappa_M^g
    &=
    a' + \frac{a'}{s} \prn*{\frac{1}{a} + a'}
    =
    \frac{a'}{s} \prn*{\frac{1}{a} + a' + s}
    \com\quad
    \kappa_N^g
    =
    \frac{1}{a'} + a + s' + \frac{a'}{s} \prn*{\frac{1}{a'} + a + s'}
    =
    \prn*{1 + \frac{a'}{s}}
    \prn*{\frac{1}{a'} + a + s'}
    \per
    \n
  \end{split}
\end{equation}
Then, we will find $(a, a', s, s')$ by considering the following optimization problem:
\begin{equation}
  \min_{(a,a',s,s') \colon a > 0, a' > 0, s > 0, s' > 0} \max\set*{\kappa_M^f(a, a', s, s'),\, \kappa_N^f(a, a', s, s'),\, \kappa_M^g(a, a', s, s'),\, \kappa_N^g(a, a', s, s')}
  \com
  \label{eq:min_coeff_indiv}
\end{equation}
where we note that since $\kappa_M^f, \kappa_N^f$ and $\kappa_M^g, \kappa_N^g$ contain $s'$ and $s$ in their denominators, respectively, it suffices to consider the case of $s > 0$ and $s' > 0$.

It suffices to consider the case of $a = a'$, $s = s'$ to find the optimal $(a, a', s, s')$ in \Cref{eq:min_coeff_indiv}.
This can be verified from the fact that, under the change of variables $p = \log a$, $p' = \log a'$, $q = \log s$, and $q' = \log s'$, the functions $\kappa_M^f, \kappa_N^f, \kappa_M^g, \kappa_N^g$ are convex in $(p, p', q, q')$  (as discussed in detail in \Cref{app:convexify}), together with the invariance of the objective in \Cref{eq:min_coeff_indiv} under the swaps $a \leftrightarrow a'$ and $s \leftrightarrow s'$.
When $a = a'$ and $s = s'$, we have
\begin{equation}
  \kappa_M^f = \kappa_N^g
  =
  \prn*{1 + \frac{a}{s}} \prn*{\frac{1}{a} + a + s}
  \eqqcolon 
  \kappa_1
  \com\quad
  \kappa_N^f = \kappa_M^g
  =
  \frac{a}{s} \prn*{\frac{1}{a} + a + s}
  \eqqcolon
  \kappa_2
  \per
  \n
\end{equation}
From this we have $\kappa_1 \geq \kappa_2$, and thus
\begin{equation}
  \max\set*{\kappa_M^f, \kappa_N^f, \kappa_M^g, \kappa_N^g}
  =
  \kappa_1
  \per
  \n
\end{equation}
The optimal values of $(a, s)$ that minimize $\kappa_1(a,s)$ is $(a, s) = (1/\sqrt{3}, 2/\sqrt{3})$,
and at that point we have $\kappa_1(a, s) = (3/2) \cdot (\sqrt{3} + 1/\sqrt{3} + 2/\sqrt{3}) = 3 \sqrt{3}$.
Consequently, the optimal argument of the optimization problem in \Cref{eq:min_coeff_indiv} is given by $(a, a', s, s') = (1/\sqrt{3}, 1/\sqrt{3}, 2/\sqrt{3}, 2/\sqrt{3})$.
Therefore, from \Cref{eq:a_s_to_c_eta}, the corresponding $(\eta, \eta', c, c')$ equals $\prn[\big]{\frac{1}{2\sqrt{3}}, \frac{1}{2\sqrt{3}}, \frac12, \frac12}$.
This completes the proof.
\end{proof}

\subsubsection{Convex Reformulation and Numerical Learning-Rate Computation}\label{app:convexify}
As discussed in \Cref{lem:optimal_J_gamma}, in the cardinality-aware setting, it is difficult to obtain closed-form learning rates $\eta,\eta'>0$ that minimize the maximum of the individual regret. 
In what follows, we discuss a convex reformulation of $f$ and $g$ (for possibly given $M=\log m$ and $N=\log n$) and numerical methods for computing $\eta,\eta'>0$ that minimize either the maximum of the individual regrets or the convex sum of individual regrets $J_\gamma$ in \Cref{eq:def_Jgamma}.

Recall that a function $f \colon \R^d \to \R$ with $\dom f = \Rp^d$ is called a monomial function if there exists $c > 0$ and $a_i \in \R$ such that $f(z) = c z_1^{a_1} z_2^{a_2} \dots z_d^{a_d}$,
and a function is a posynomial if it can be written as a sum of monomials \citep[Section~4.5]{boyd04convex}. 
An important observation is that $f$ and $g$ in \Cref{eq:def_fg_arg_as} are posynomials over $(a,a',s,s')\in\R_p^4$. 
As noted in the main text, optimizing $f$ and $g$ is in general nonconvex in the original variables, but by performing the change of variables described below, one can convert the problem into a convex optimization problem and solve it efficiently (see \eg~\citealt[Section~4.5.3]{boyd04convex}) 
(As before, after the change of variables we will reuse the same symbols for the transformed functions  by abuse of notation).
We use the change of the variables given by
\begin{equation}
  p = \log a
  \com\quad
  p' = \log a'
  \com\quad
  q = \log s
  \com\quad
  q' = \log s'
  \per
  \n
\end{equation}
Then, we have
\begin{equation}
  h(p, p')
  =
  M \e^{- p}
  +
  M \e^{p'}
  +
  N \e^{-p'}
  +
  N \e^{p}
  +
  \frac12 \e^p
  +
  \frac12 \e^{p'}
  \com
  \n
\end{equation}
and thus we can rewrite $f$ and $g$ in \Cref{eq:def_fg_arg_as} as
\begin{equation}
  \begin{split}
    f(p,p',q,q')
    &= 
    M \e^{-p}
    + 
    M \e^{p'}
    + 
    \prn*{N+\frac{1}{2}} \e^{p}
    +
    M \e^{q}
    \\
    &\qquad+ 
    M \e^{-q'}
    +
    \prn*{M+\frac{1}{2}} \e^{p+p'-q'}
    + 
    N \e^{p-p'-q'}
    + 
    \prn*{N+\frac{1}{2}} \e^{2p-q'}
    + 
    M \e^{p+q-q'}
    \com
  \end{split}
  \n
\end{equation}
and
\begin{equation}
  \begin{split}
    g(p,p',q,q')
    &=
    N \e^{-p'}
    +
    N \e^{p}
    +
    \prn*{M+\frac{1}{2}} \e^{p'}
    +
    N \e^{q'}
    \\
    &\qquad+
    N \e^{-q}
    +
    \prn*{M+\frac{1}{2}} \e^{2p'-q}
    +
    \prn*{N+\frac{1}{2}} \e^{p+p'-q}
    +
    M \e^{p'-p-q}
    +
    N \e^{p'+q'-q}
    \per
  \end{split}
  \n
\end{equation}
Since $f$ and $g$ are convex in $(p,p',q,q')$, both their pointwise maximum and any convex combination are also convex in $(p,p',q,q')$. 
Hence their (globally optimal) solutions can be computed efficiently. 
In the cardinality-aware case, several propositions in the main text selected slightly compromised learning rates in order to admit closed-form expressions. 
However, if one is allowed to solve the above convex optimization, one can numerically obtain learning rates that further minimize the maximum of the individual regrets or the convex sum of individual regrets. 
We exploited this convexification to produce \Cref{fig:indiv_reg_vs_gamma}.

\section{Omitted Details from \Cref{sec:lower_bound}}\label{app:proof_lower_bound}
Here we provide the deferred details from \Cref{sec:lower_bound}.

\begin{proof}[Proof of \Cref{lem:ineq_telescope}]
We first show that the function $f(a) = \log(1 + z \e^{-a})$ is convex with respect to~$a$ for $z > 0$.
This can be confirmed from the fact that 
\begin{equation}
  f''(a)
  =
  \frac{- z \e^{-a}}{1 + z \e^{-a}}
  \com
  \quad
  f''(a)
  =
  \frac{z \e^{-a}}{\prn{1 + z \e^{-a}}^2}
  >
  0
  \per
  \nn
\end{equation}
Hence, from the convexity of $f$, we have
$f(a) - f(0) \geq f'(0) \cdot b$, which is equivalent to
\begin{equation}
  \log(1 + z \e^{-a})
  -
  \log(1 + z)
  \geq
  \frac{- z}{1 + z} \cdot a
  \per
  \n
\end{equation}
Rearranging the last inequality completes the proof.
\end{proof}
\section{Omitted Details from \Cref{sec:dynamic}}\label{app:proof_dynamic}
This section provides the proof of \Cref{thm:lower_bounds_dynamic}.

\begin{proof}[Proof of \Cref{thm:lower_bounds_dynamic}]
We consider the game with the payoff matrix $A$ in \Cref{eq:def_A}.
Then, 
noting that the optimal strategy for each player is to keep choosing only action 1 (that is to play the pure strategy $e_1$),
and following the analysis used in the proof of \Cref{thm:lower_bounds}, we can evaluate the dynamic regret as
\begin{equation}
  \DReg^x_T
  =
  \sumT \Delta (1 - x_t(1))
  \com
  \label{eq:dynreg_for_A}
\end{equation}
where we note that $x_t$'s are the time-averaged strategy of the optimistic Hedge as given by \Cref{eq:def_average_iter}.

Following the analysis used in the proof of \Cref{thm:lower_bounds} again, for each $t \in [T]$ we have
\begin{equation}
  \xhat_t(1)
  =
  \frac{1}{
    1
    +
    \alpha_t
  }
  \com
  \quad
  \alpha_t
  =
  (m-1)
  \exp\prn*{
    - \eta \Delta t
  }
  \per
  \label{eq:xhattone_for_A}
\end{equation}
Combining \Cref{eq:dynreg_for_A,eq:xhattone_for_A}, we can lower bound the dynamic regret of the $x$-player as
\begin{equation}
  \DReg_T^x
  =
  \Delta
  \sumT 
  \prn*{1 - \frac{1}{t} \sum_{s=1}^t \frac{1}{1 + \alpha_s}}
  =
  \Delta
  \sumT 
  \frac{1}{t}
  \sum_{s=1}^t 
  \frac{\alpha_t}{1 + \alpha_s}
  =
  \Delta
  \sum_{s=1}^T
  \frac{\alpha_s}{1 + \alpha_s}
  \sum_{t=s}^T
  \frac{1}{t}
  \com
  \label{eq:dynreg_equality}
\end{equation}
where in the last inequality, we exchanged the order of summation.
For any $S_0 \in [T]$, 
the last quantity is lower bounded as
\begin{align}
  \Delta
  \sum_{s=1}^T
  \frac{\alpha_s}{1 + \alpha_s}
  \sum_{t=s}^T
  \frac{1}{t}
  &\geq
  \Delta
  \sum_{s=1}^{S_0}
  \frac{\alpha_s}{1 + \alpha_s}
  \sum_{t=s}^T
  \frac{1}{t}
  \nn
  &\geq
  \Delta
  \sum_{s=1}^{S_0}
  \frac{\alpha_s}{1 + \alpha_s}
  \log \prn*{\frac{T + 1}{s}}
  \nn
  &\geq
  \Delta
  \log \prn*{\frac{T + 1}{S_0}}
  \sum_{s=1}^{S_0}
  \frac{\alpha_s}{1 + \alpha_s}
  \com
  \label{eq:twosums_lower}
\end{align}
where the second inequality follows from
$\sum_{t=s}^T 1/t \geq \int_{t=s}^{T+1} (1/z) \d z = \log((T+1) / s)$ for any $s \in [T]$.

From \Cref{lem:ineq_telescope} with $z = \alpha_s$ and $a = \eta \Delta$, we have
\begin{equation}
  \frac{\alpha_s}{1 + \alpha_s}
  \geq
  \frac{1}{\eta \Delta}
  \prn*{
    \log (1 + \alpha_s)
    -
    \log (1 + \alpha_{s+1})
  }
  \com
  \n
\end{equation}
and thus
\begin{align}
  \sum_{s=1}^{S_0}
  \frac{\alpha_s}{1 + \alpha_s}
  &\geq
  \frac{1}{\eta \Delta}
  \prn*{
    \log (1 + \alpha_1)
    -
    \log (1 + \alpha_{S_0+1})
  }
  \nn
  &\geq
  \frac{1}{\eta \Delta}
  \prn*{
    \log m
    -
    \eta \Delta
    -
    (m - 1) \exp\prn{ - \eta \Delta (S_0 + 1) }
  }
  \com
  \label{eq:alphasum_lower}
\end{align}
where in the last inequality we used
$\log(1 + \alpha_1) = \log (1 + (m - 1) \exp(- \eta \Delta)) \geq \log (m \exp(- \eta \Delta))$
and $\log(1 + z) \leq z$ for $z \in \R$.

Combining \Cref{eq:dynreg_equality,eq:twosums_lower,eq:alphasum_lower},
we can lower bound the dynamic regret of the $x$-player as
\begin{align}
  \DReg_T^x
  &
  \geq
  \frac{1}{\eta}
  \log \prn*{\frac{T + 1}{S_0}}
  \prn*{
    \log m
    -
    \eta \Delta
    -
    (m - 1) \exp\prn{ - \eta \Delta (S_0 + 1) }
  }
  \per
  \n
\end{align}
Since $S_0 \in [T]$ is arbitrary, choosing $S_0 = \sqrt{T + 1}$ gives
\begin{equation}
  \DReg_T^x
  \geq
  \frac{1}{2 \eta}
  \log \prn*{T + 1}
  \prn[\big]{
    \log m
    -
    \eta \Delta
    -
    (m - 1) \exp\prn{ - \eta \Delta (\sqrt{T+1} + 1) }
  }
  \per
  \n
\end{equation}
Finally, choosing
\begin{equation}
  \Delta
  =
  \min\set*{
    1,
    \frac{\log \prn{(m - 1)(\sqrt{T + 1}+ 1)}}{\eta (\sqrt{T + 1} + 1)}
  }
  \n
\end{equation}
in the last inequality as in the proof of \Cref{thm:lower_bounds},
we obtain the desired lower bound for the $x$-player.
The dynamic regret of the $y$-player can be lower bounded by the same argument, and we have completed the proof.
\end{proof}

\section{Numerical Experiments}\label{app:experiments}
In this section, we present numerical experiments comparing the performance of the eight learning dynamics investigated in this paper.
We then demonstrate that, in settings where there is a gap between the sizes of $m$ and $n$, as discussed in \Cref{sec:introduction}, being cardinality-aware learning dynamics indeed improves the performance of corresponding cardinality-unaware learning dynamics.

\paragraph{Setup}
The eight learning dynamics used for comparison in the experiments are summarized in \Cref{table:lr}.
This table is obtained by modifying \Cref{table:regret} to remove the lower-bound entries and to summarize the optimal learning rates $\eta, \eta'$ for each target regret (\ie~social, individual, or the maximum of the individual regrets).
For the performance comparison, we used the payoff matrix defined in \Cref{eq:def_A} with $\Delta = 1$.
We set the numbers of actions to $(m, n) = (2, 10^4)$ so that their difference is large, and the number of rounds to $T = 2000$.
We evaluated four metrics: the social regret $\Reg_T^x + \Reg_T^y$, the maximum of the individual regrets $\max\set{\Reg_T^x, \Reg_T^y}$, the $x$-player's regret $\Reg_T^x$, and the $y$-player's regret $\Reg_T^y$.

\begin{figure}[t]
  \centering
  \begin{minipage}[t]{0.485\linewidth}
    \centering
    \includegraphics[width=\linewidth]{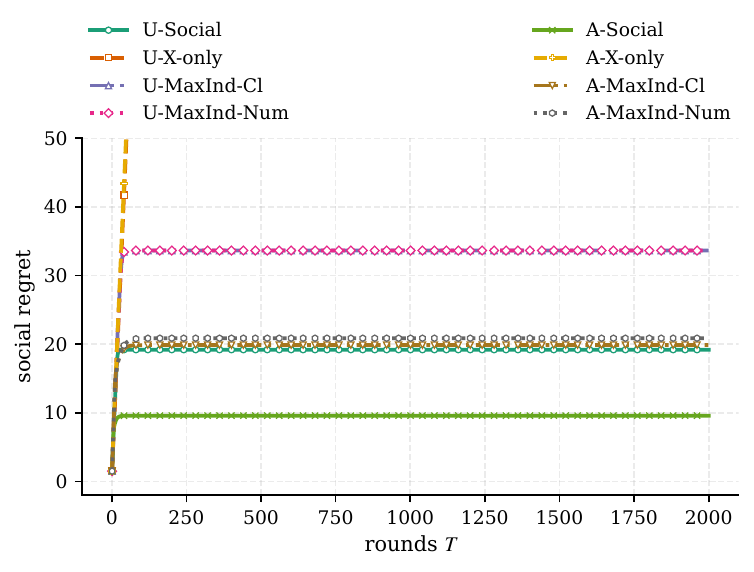}
  \end{minipage}\hfill
  \begin{minipage}[t]{0.485\linewidth}
    \centering
    \includegraphics[width=\linewidth]{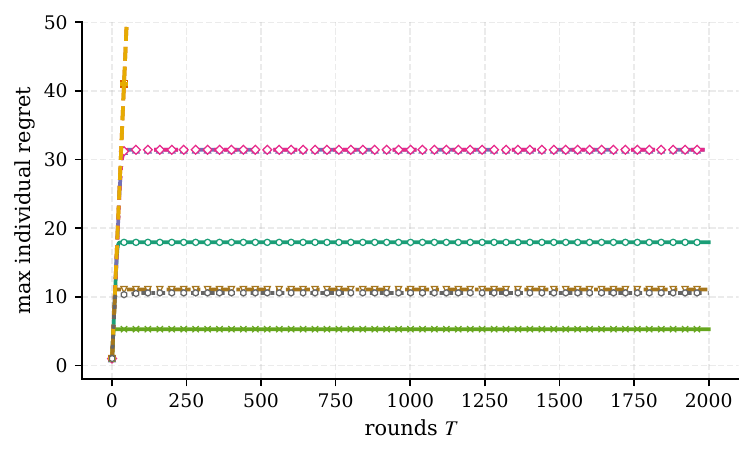}
  \end{minipage}\hfill
  \vspace{5pt}
  \\
  \begin{minipage}[t]{0.485\linewidth}
    \centering
    \includegraphics[width=\linewidth]{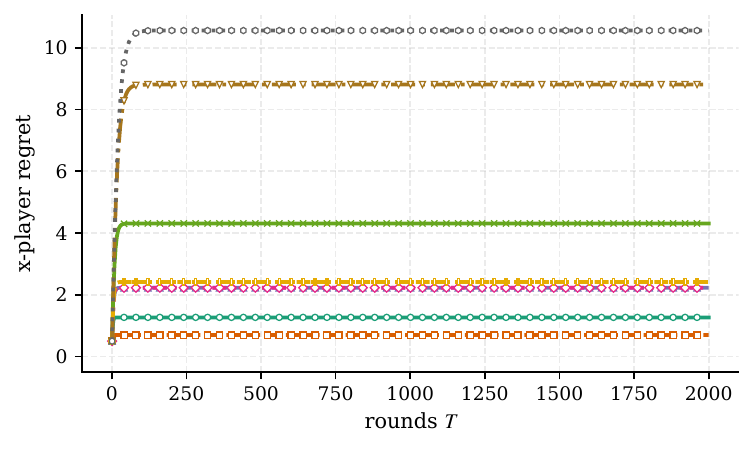}
  \end{minipage}\hfill
  \begin{minipage}[t]{0.485\linewidth}
    \centering
    \includegraphics[width=\linewidth]{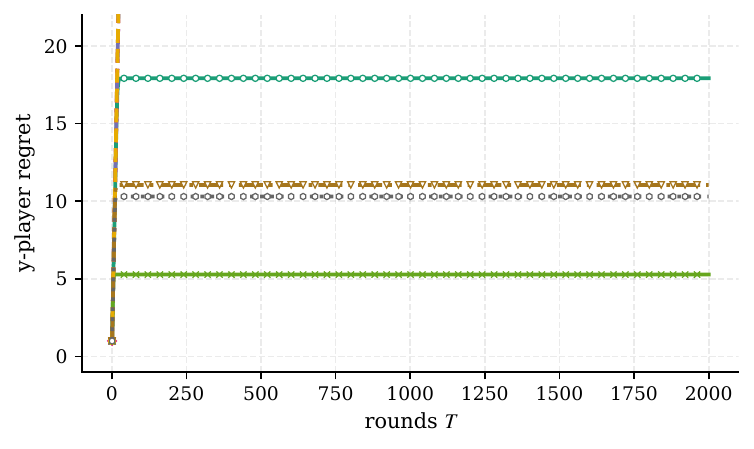}
  \end{minipage}
  \caption{Regret versus the number of rounds for the learning dynamics based on the optimistic Hedge algorithm in the setting $m = 2$ and $n = 10^4$.
  From top-left to bottom-right: social regret, the maximum of the individual regrets, the regret of the $x$-player, and the regret of the $y$-player.
  }
  \label{fig:m2n10000}
\end{figure}

\paragraph{Results}
The results are provided in \Cref{fig:m2n10000}.
For the social regret, the learning dynamic that minimizes the social regret under the cardinality-aware setting (\texttt{A-Social}) achieves the best performance, significantly improving upon the best cardinality-unaware algorithm (\texttt{U-Social}).
For the maximum of the individual regrets, \texttt{A-Social} again achieves the best performance, followed by the approaches that directly minimize the maximum of the individual regrets under the cardinality-aware setting (\texttt{A-MaxInd-Cl} and \texttt{A-MaxInd-Num}).
These results demonstrate that incorporating cardinality-awareness leads to clear performance improvements over the cardinality-unaware case.

It should be noted that \texttt{A-Social}, which achieved the best performance in terms of both social regret and the maximum of the individual regrets, is not theoretically guaranteed to upper bound the individual regret, as discussed in the main text.
This suggests that the algorithm may have empirically achieved low individual regret under this particular instance, or that we can upper bound the individual regrets under this choice of learning rates.
As discussed in \Cref{sec:conclusion}, a more detailed investigation of this remains an important direction for future work.

\begin{table*}[t]
    \caption{
      Eight learning dynamics compared in the numerical experiments and their learning rates.
      Here, $M = \log m$, $N = \log n$, $M' = M + 1/2$, $N' = N + 1/2$, and $D = \sqrt{M' N'} + \sqrt{M N}$.
    }
    \label{table:lr}
    \centering
    \resizebox{\textwidth}{!}{
    \footnotesize
    \begin{tabular}{lllll}
      \toprule
      Name & Target & Upper bound & Proposition & Learning rate
      \\
      \midrule
      \multicolumn{4}{l}{{\normalsize $\triangleright$ \uline{Strongly-uncoupled learning dynamics (cardinality-unaware)} }} & 
      \vspace{2pt}
      \\
     \texttt{U-Social} & Social regret & $2 (M + N) + 1$ & \Cref{thm:optimal_sreg_su} & $\eta=1/2$, $\eta'=1/2$
      \\
      \texttt{U-X-only} &  $x$-player's regret & $M + N + 1/2$ & \Cref{eq:x_best_indiv_su} & $\eta = 1$, $\eta' = 0$
      \\
      \texttt{U-MaxInd-Cl} & Max of indiv.~regrets & $3 \sqrt{3} (M + N) + {1}/{\sqrt{3}}$ & \Cref{thm:optimal_max_indivreg_su} & $\eta=1/(2 \sqrt{3})$, $\eta'=1/(2\sqrt{3})$
      \\
      \texttt{U-MaxInd-Num} & Max of indiv.~regrets & same as above & --  & Numerically computed (\Cref{app:convexify})
      \\
      \midrule
      \multicolumn{4}{l}{{\normalsize $\triangleright$ \uline{Cardinality-aware strongly-uncoupled learning dynamics}}} & 
      \vspace{2pt}
      \\
      \texttt{A-Social} & Social regret & $2 \sqrt{M N'} + 2 \sqrt{M' N}$ & \Cref{thm:optimal_sreg_casu} & 
      $\eta=\frac{\sqrt{M M'}}{D}$, $\eta'=\frac{\sqrt{N N'}}{D}$ 
      \\
      \texttt{A-X-only} & $x$-player's regret & $2 \sqrt{M N'}$  & \Cref{eq:x_best_indiv_casu} & $\eta = \sqrt{M / N'}$, $\eta' = 0$
      \\
      \texttt{A-MaxInd-Cl} & Max of indiv.~regrets & $(20/3) \prn{ \sqrt{M N'} + \sqrt{M' N} }$ & \Cref{thm:optimal_max_indivreg_casu} & $\eta=\frac{\sqrt{M M'}}{2 D}$, $\eta'=\frac{\sqrt{N N'}}{2 D}$      \\
      \texttt{A-MaxInd-Num} & Max of indiv.~regrets & $C \prn{ \sqrt{M N'} + \sqrt{M' N} } \ (C < \frac{20}{3})$ & --  & Numerically computed (see \Cref{app:convexify})
      \\
      \bottomrule
    \end{tabular}
    }
  \end{table*}

\end{document}